\pdfoutput=1
\documentclass[sigconf]{acmart}

\usepackage{booktabs} 
\usepackage{float}
\usepackage{longtable}
\usepackage{subcaption}
\usepackage{multirow}
\usepackage{tikz}
\graphicspath{{figures/}}

\copyrightyear{2019}
\acmYear{2019}
\setcopyright{acmlicensed}
\acmConference[IUI '19]{24th International Conference on Intelligent User Interfaces}{March 17--20, 2019}{Marina del Rey, CA, USA}
\acmBooktitle{24th International Conference on Intelligent User Interfaces (IUI '19), March 17--20, 2019, Marina del Rey, CA, USA}
\acmPrice{15.00}
\acmDOI{10.1145/3301275.3302265}
\acmISBN{978-1-4503-6272-6/19/03}

\begin{document}
\title{What can AI do for me?}
\subtitle{Evaluating Machine Learning Interpretations in Cooperative Play}

\author{Shi Feng}
\affiliation{%
  \institution{University of Maryland}
  \city{College Park}
  \state{Maryland}
}
\email{shifeng@cs.umd.edu}

\author{Jordan Boyd-Graber}
\affiliation{%
  \institution{University of Maryland}
  \city{College Park}
  \state{Maryland}
}
\email{jbg@umiacs.umd.edu}

\newcommand\BibTeX{B{\sc ib}\TeX}
\newcommand{\qb}[0]{Quizbowl}
\newcommand{\abr}[1]{\textsc{#1}}
\newcommand{\vis}[1]{\emph{#1}}
\newcommand{\etal}[0]{\emph{et al.}}
\definecolor{lightblue}{HTML}{3cc7ea}
\definecolor{periwinkle}{rgb}{0.8, 0.8, 1.0}
\definecolor{colorsquad}{rgb}{0,1,0}
\definecolor{colorsnli}{rgb}{1,0,0}
\definecolor{colorvqa}{rgb}{1,1,0}
\newcommand{\ai}[0]{\textsc{ai}}

\newif\ifcomment\commentfalse
\ifcomment
 \newcommand{\jbgcomment}[1]{ \colorbox{red}{   \parbox{.8\linewidth}{ JBG: #1}  }}
 \newcommand{\fscomment}[1]{ \colorbox{green}{   \parbox{.8\linewidth}{ fs: #1}  }}
\else
\newcommand{\jbgcomment}[1]{}
\newcommand{\fscomment}[1]{}
\fi

\begin{abstract}
Machine learning is an important tool for decision making, but its
ethical and responsible application requires rigorous vetting of its
interpretability and utility: an understudied problem, particularly
for natural language processing models.  
We propose an evaluation of interpretation on a
real task with real human users, where the effectiveness of
interpretation is measured by how much it improves human performance.
We design a grounded, realistic human-computer cooperative setting
using a question answering task, \qb{}.
We recruit both trivia experts and novices to play this game with
computer as their teammate, who communicates its prediction via three
different interpretations.
We also provide design guidance for natural language processing
human-in-the-loop settings.

\end{abstract}

%
%

\ccsdesc[500]{Human-centered computing~Natural language interfaces}
\ccsdesc[500]{Human-centered computing~Collaborative interaction}

\keywords{interpretability; natural language processing; question answering}

\maketitle

\section{Introduction}
\label{sec:intro}

The field of machine learning (\abr{ml}) is making rapid progress,
with models surpassing human performance on many tasks, such as
image classification~\cite{he2015delving}, playing video
games~\cite{mnih2015human}, and playing Go~\cite{silver2017mastering}.
However, a drop-in replacement for humans---even assuming that it is
achievable---is not always the ideal integration of machine learning
into real-world decision making.
In sensitive areas such as medicine and criminal justice, the
computational objectives of \abr{ml} models cannot yet fully
capture the factors one must consider when making a
decision, such as fairness and transparency. In some other areas such
as natural language processing, the
strengths of humans and computers are sometimes complimentary.
Humans are excellent at reasoning about what we
consider ``common sense'', while some tasks in this category such as
disambiguating word senses are still difficult for
computers~\cite{papandrea2017supwsd}.
Tasks like deceptive review detection is difficult and time consuming
for humans while simple linear \abr{ml} models achieve high accuracy
with little processing time~\cite{lai2018human}.
On tasks such as simultaneous interpretation where humans are
still far superior than computers, experts can still be assisted on
some aspects of the task: interpreters often
find certain content such as technical terms, names of people and
organizations, and numbers difficult to translate, while 
computers find that easy.  The integration of \abr{ml}
can be more effective and efficient when humans and computers
cooperate.

Cooperation is only effective when the two parties communicate
well with each other. One direction of this communication, from
humans to computers, is well-studied:
\abr{ml} models can be improved with human feedback using
reinforcement learning~\cite{sutton1998introduction} and imitation
learning~\cite{ross2011reduction,ross2018regularizing}. The
other direction of the communication, from \abr{ml} models to
humans, presents different challenges:
a standard classification model outputs a prediction (e.g., an object
class given an image), but without any justification.
Although the prediction can be presented with a confidence score (a
value between zero and one), humans struggle to interpret and act on
numbers~\cite{peters2006numeracy, reyna2008numeracy}; moreover, due to
over-fitting, confidence scores from a neural models can be much
higher than the actual prediction
uncertainty~\cite{guo2017calibration}.

To bridge the gap between human and \abr{ml} models in a cooperative
setting, interpretation methods 
explain the model predictions in a more expressive,
human-intelligible way. In a human-centered setting where humans make
the final decision, these methods help users decide to trust the
model prediction or not. In Section~\ref{sec:related} we discuss the
existing work of interpreting \abr{ml} models.

Progress in \abr{ml} research largely relies on rigorous evaluations,
which often relies on standard
datasets, for example ImageNet~\cite{deng2009imagenet} for image
classification and Penn Treebank~\cite{marcus1993building} for
language modeling.
Although interpretability is valued as a laudable goal, it remains
elusive to evaluate.
We do not have such standard dataset for interpretability---it is not
clear what the ground truth should be. As
Lipton~\cite{lipton2016mythos} argues, there is no clear agreement on what
interpretability means; there is no definitive answer to what
interpretation is most faithful to the model and useful for humans at
the same time. Secondly, it is not realistic to evaluate
interpretability without humans, the eventual consumer of
interpretations~\cite{narayanan2018humans}.
Previous work focuses on how humans can use interpretations to help
the model do its job better; for example, interpretations generated by
Local Interpretable Model-Agnostic
Explanations~\cite[\abr{lime}]{ribeiro2016lime} help humans do feature
engineering to improve downstream predictions of a classifier; in
other work interpretations are used to help humans debug \abr{ml}
models~\cite{ribeiro2018semantically,fong2017interpretable}.

Kleinberg~\etal{}~\cite{kleinberg2017human} propose a different
perspective and ask how \abr{ml} can improve human decision making.
Applying this thinking, we measure
interpretability by asking what \abr{ml} can do
for humans through interpretations: they should
\emph{augment}~\cite{Koedinger-13} human intelligence. This concept
resonates with the seminal work of mixed-initiative user
interface~\cite{horvitz1999principles}, which emphasizes 
user interfaces where the human and the computer can drive towards a
shared goal and ones that enhance human ability~\cite{allen1999mixed}.

Interpretations come in many forms; we focus on three
popular options among the interpretable \abr{ml} community:
visualizing uncertainty, highlighting
important input features, and retrieving relevant training
examples.  We measure how they help humans on the tasks at hand and
focus on answering the question ``how effective can interpretations
communicate model predictions to humans''. The other question is ``how
faithful an interpretation is to the model''.
Section~\ref{sec:eval_qb} discusses our choice of model to answer the
first question; we leave the second question to future work, but
discuss in Section~\ref{sec:discussion} how our framework, interface,
and experiments can be directly applied.

We choose the testbed for our interpretability evaluation from the
natural language domain---a question answering task called
\qb{}~\cite{boydgraber2012besting}.  As we discuss in
Section~\ref{sec:eval_qb}, in addition to being a
challenging task for \abr{ml}, it is also an exciting game that is
loved by human trivia enthusiasts. Furthermore, it is a task where
humans and \abr{ml} have complementary strengths, so effective
collaboration with interpretations has great potential.

We recruit both \qb{} enthusiasts and turkers from Amazon Mechanical
Turk (novices in comparison) to play \qb{} on an interactive
interface,
provide them different combinations of the interpretations, and
measure how their performance changes.  These different user groups
reveal imperfections in how we communicate the way a computer answers
questions.  Experts have enough world and task expertise to
confidently overrule when the computer is wrong; however, as we will
discuss in Section~\ref{sec:results}, novices are too trusting: they
play more aggressively with computer assistance, but are not able to
discern useful help from the misleading ones as well as the experts.
In Section~\ref{sec:discussion}, we propose how to can explore new
interpretations and visualizations to help humans more confidently
interpret \abr{ml} algorithms.

\section{Related Work}
\label{sec:related}

\subsection{Human-AI Cooperation}
Explainability is a central problem of applied \abr{ai}, with research
stretching back to the days of expert
systems~\cite{swartout1983xplain}.
The recent surge of interest in this area is the
result of the success of \abr{ml} models based on neural networks,
a.k.a.\ deep learning~\cite{lecun2015deep}. These complicated models
have stupendous predictive power, but at the same time brittle, best
demonstrated by the existence of adversarial
examples~\cite{goodfellow2014explaining}, where small perturbation to
the input leads to significant change in the model output.
From a practical standpoint, the inscrutability of these models makes
it difficult to integrate into real world decision-making in high risk
areas such as urban planning, disease diagnosis, predicting insurance
risk, and criminal justice.
The fairness, accountability, and transparency of machine learning
remain a concern~\cite{acm2017public}, which is reflected in the
``right to explanation'' in European Union's new General Data
Protection Regulation~\cite[\abr{gdpr}]{gdpr}. 

Thus, \abr{ml} model predictions need explanations.
Efforts including the Explainable \abr{ai} (\abr{xai})
initiative~\cite{gunning2017explainable} led to the conceptualization
of a series of human-\abr{ai} cooperation paradigms, including
human-aware \abr{ai}~\cite{chakraborti2017ai}, and human-robot
teaming~\cite{vinson2018human}. 
As an example, Schmidt and Herrmann~\cite{schmidt2017intervention}
recognize the importance of interpretability when interacting with
autonomous vehicles.
Such need motivated the \abr{ml} community to develop interpretation
methods for deep neural models~\cite[\em inter
alia]{baehrens2010explain, simonyan2013deep}.


The \abr{hci} community has a rich body of research towards making
computers more usable, for example in interaction
design~\cite{ju2008design} and software
learnability~\cite{grossman2009survey}.
To borrow insights from the human side,
Miller~\cite{miller2017explanation} provides an overview of social
science research regarding how people define, generate, select,
evaluate, and present explanations.
Still, interpreting \abr{ml} models has its unique challenges.
Krause~\etal{}~\cite{krause2016interacting} compare different \abr{ml}
models under one visualization method, partial dependency.
Smith~\etal{}~\cite{smith2017evaluating} and
Lee~\etal{}~\cite{lee2017human} focus on the
interpretation of topic models. In contrast, we compare interpretation
of classification models across various forms, making
our framework more generalizable to other tasks and interpretation
methods.

\subsection{Interpretation of Machine Learning Models}

Interpretations can take on several different forms.
We focus on interpretation in the form of uncertainty, important
input features, and relevant training examples.  Some \abr{ml} models
provide canonical interpretations.
For models such as
decision trees and association rule
lists~\cite{lakkaraju2016interpretable, letham2015interpretable}, the
interpretation is built in the prediction itself.  However, most
state-of-the-art models in vision and language---domains with the
widest range of applications---are deep neural models with hundreds of
thousands of parameters. Next we introduce previous work on
interpreting both simpler linear models and more complicated neural
networks, in each of the three forms.

\paragraph{Conveying Uncertainty}
Augmenting the prediction from a neural network classifier with a
confidence score (a value between zero and one) conveys the
uncertainty of the model. In a cooperative setting, the uncertainty
helps humans decide to trust the model or
not~\cite{antifakos2005towards, rukzio2006visualization}. To make it
more informative, we can also display the confidence for the
classes other than the top one~\cite{liu2017towards}.
Confidence of simple linear models are usually
well-calibrated, but estimating uncertainty for a deep neural model
is challenging: due to overfitting, they are over-confident 
and require careful calibration~\cite{guo2017calibration,
feng2018rawr}.

\paragraph{Highlighting Important Features}
Model predictions can be explained by
highlighting the most salient features in the input, typically visualized
by a heat map.  For a linear classifier, the most salient features
are the ones with the largest corresponding coefficients;  For
non-linear classifiers, the relevance of a feature can be calculated
by the gradient of the loss function w.r.t.\ that
feature~\cite{simonyan2013deep}.
Alternatively, one can locally approximate a non-linear classifier
with a simpler linear model, then use the coefficients to explain the
predictions from the non-linear model~\cite{ribeiro2016lime}.

\paragraph{Interpretation by Example}
We can explain a prediction on a test example by finding the most
influential training examples.  Various metrics exist for finding
important training examples, such as distance in the representation
space which is natural to linear models, clustering algorithms and
their deep variation~\cite{papernot2018dknn}, and influence
functions~\cite{koh2017influence} for non-linear models.

As we discuss in Section~\ref{sec:eval_qb}, although our experiments
use a linear classifier,  our method can be generalized to evaluating
these methods designed for neural models
(Section~\ref{sec:discussion}).

\subsection{Evaluation of Interpretation}

A fair and accurate assessment of interpretations is crucial for
improving the understability of \abr{ai} and consequently
human-\abr{ai} cooperation.
Although interpretation methods have rigorous mathematical
formulations, some even axiomatically
derived~\cite{sundararajan2017axiomatic}, it remains unclear how we
can evaluate the efficacy of these methods on \emph{real tasks} with
\emph{real users}. Lipton~\cite{lipton2016mythos} argues that there is no
clear agreement on what interpretability means: looking at \abr{ml}
models alone, no definitive answer exists as in what would be the best
interpretation in both faithfulness to the model and usefulness
to humans.

As it is widely accepted that machine learning models should be
evaluated beyond natural examples, e.g., in adversarial
settings~\cite{goodfellow2014explaining, jia2017adversarial},
the evaluation of interpretation should not be limited to being
visually pleasing. Indeed, interpretations can be fragile under small
input perturbations~\cite{ghorbani2017interpretation,
kindermans2017unreliability}, unfaithful to the
model~\cite{hooker2018evaluating, adebayo2018sanity, feng2018rawr},
and create a false sense of security~\cite{jiang2018trust}.

Conditioning a more realistic setting, Doshi-Velez and
Kim~\cite{doshivelez2017towards}
provide an ontology of various evaluations of interpretation with a
human in the loop. Following this framework,
Narayanan~\etal{}~\cite{narayanan2018humans} conduct
one such evaluation with synthetic tasks and hand-crafted
interpretations to study their desirable cognitive properties.

We focus on \emph{application-grounded} evaluation---real tasks with
real users.  This setting best aligns with what interpretations are
intended for---improving human performance on the end task.  However,
application-grounded evaluation is also challenging because it
requires real tasks and motivated real users. The task needs a large
pool of willing human testers, and ideally one that challenges both
humans and computers. As we discuss in the next section, \qb{} is a
task that satisfies these conditions.

\section{Interpretation Testbed: Quizbowl}
\label{sec:eval_qb}

This section introduces \qb{}, our testbed for evaluating the three
forms of interpretations. We discuss how the task suits our purposes,
which model to use, and how we generate the interpretations.

\subsection{\qb{} and Computer Models}

\qb{} is both a challenging task for
machine learning~\cite{boydgraber2012besting} and a trivia game played
by thousands of students around the world each year. Each question
consists of multiple clues, presented to the players
\emph{word-by-word}, verbally or in text.
The ordering of \qb{} clues is
\emph{pyramidal}---difficult clues at the beginning, easy clues at the
end, and the challenge is to answer with as few clues as possible.
For a question with $n$ words, the players have $n$ chances to decide
that \emph{this is all the information I need to answer the question}.
The player can do so by \emph{buzzing} before the question is fully
read, which interrupts the readout so
the player provide an answer. Whoever gets the answer correct first
wins that question and receives ten points.\footnote{Like previous
work, we only consider \emph{toss-up}/\emph{starter} questions.} But
when players buzz and answer incorrectly, they lose five points.
Success in \qb{} requires a player to not only be knowledgeable but
also balance between aggressiveness and
accuracy~\cite{he2016opponent}.

\qb{} challenges humans and computers in different
ways~\cite{boydgraber2012besting, wallace2018trick}.  Computers can
memorize every poem and book ever written, making it trivial to
identify quotes.  Computers can also memorize all of the \emph{reflex
clues} that point to answers (e.g., if you hear ``phosphonium ylide'',
answer \underline{Wittig}) and apply them without any higher reasoning.
Humans can chain together evidence (``predecessor of the Queen who
pardoned Alan Turing'') and solve wordplay (``opera about an enchanted
woodwind instrument'').  
Thus, \qb{} is representative of tasks where human-computer
cooperation 
has huge potential~\cite{Thompson-13}. This also makes \qb{} a
suitable testbed for interpretation methods designed to better
interface humans and computers.

Thus, instead of trying to beat humans with computers, we team
them together and use their cooperation to
measure the effectiveness of interpretations.  In our cooperative
setting, instead of having a model to decide when to buzz in,
\emph{the human needs to decide when the system has a good guess}.
When answering a \qb{} question---which takes many steps, the human
constantly interacts with the model, which provides many opportunities
to evaluate the interpretability of models.  Every word provides new
evidence that can change the underlying interpretation and convince
the human that the system has a good answer to offer.
Furthermore, the competitiveness of
\qb{} encourages humans to use the help from the computers as much as
possible, avoiding a degenerate scenario where the users solve the
task on their own. It also attracts a large pool of enthusiastic
participants, which is crucial for application-grounded evaluations.
Sesction~\ref{sec:setup} discusses the cooperation in detail.


As mentioned in Section~\ref{sec:intro}, we focus on the comparison
between three forms of interpretation, using one method for each form.
But which method to use?  Linear models provide canonical
interpretations: important features and relevant training examples can
be identified based on the coefficients. On the other hand, neural
models do not have
canonical interpretations: 
all interpretations are approximations, which by definition are not
completely faithful to the model~\cite{rudin2018please}.

Luckily in the case of \qb{}, we have linear models with
performance on par or better than neural models.
\abr{qanta}~\cite{iyyer2014ann} is a simple, powerful, and
interpretable system for \qb{}. A stripped-down, minimal version of it
is provided to participants in the \abr{nips} 2017 Human-Computer
Question Answering competition~\cite{nips2018qbcomp}.  We use the
\emph{guesser} of \abr{qanta}, which has a linear decision function
built on ElasticSearch~\cite[\abr{es}]{gormley2015es}.  As the name
implies, guesser generates guesses for what the answer to a question
could be.  Despite its simplicity, \abr{es}-based
systems perform very well on \qb{}, defeating top trivia
players.\footnote{\url{https://youtu.be/bYFqMINXayc}}

\subsection{Interpretation of a Question Answering Model}


Our goal is to see which forms of interpretation are most
helpful to the users, and a linear model with natural
interpretations makes this easy.  Our \abr{es}-based \qb{} model
supports three forms of interpretations, each corresponding to a class
of methods widely studied in recent literature as mentioned in the
previous section.
Given a question never seen in the training set, \abr{es} mainly uses
tf-idf features to find the
most relevant training example, which is either a Wikipedia page or a
previously seen \qb{} question, and then uses the label of that
document as the answer.

To convey the uncertainty of model predictions, we augment the top ten
guesses from our model with their corresponding scores. 
Unlike regular classification models, \abr{es} does not output 
a probability distribution over all possible answers.  Its scores
measure the relevance between the question and training examples, but
are not normalized.
We keep the scores unnormalized to stay true to the model.
Despite its simplistic form, these scores provide strong
signal about model uncertainty, for example, a large gap between the
top two scores usually indicate a confident prediction.

Interpretation by example---getting the \emph{evidence}---is
straightforward with our \abr{es}-based model.
The prediction is the label of the most relevant documents, so
the extracted documents are naturally the most salient training
examples.  We can further identify the most important words in each
retrieved training example, using the highlight
\abr{api}\footnote{\url{https://www.elastic.co/guide/en/elasticsearch/reference/current/search-request-highlighting.html}}.
This gives us \emph{evidence highlights}.
The player can make a better decision of whether to trust the computer
prediction by judging how relevant the evidence is to the question.

To highlight important input features---generating \emph{question
highlights}---we build on the previous \emph{evidence highlights}.
The most
important words in the question naturally emerge when we compare the
question against the most salient training example.  Specifically, we
go through the question and find words that appear highlighted in the
evidence. Question highlights inform the player whether the computer
is looking at the right keywords in the question.

Although generating \emph{question highlights} depends on
\emph{evidence highlights}, the former can be displayed without the
later. We discuss how we control which interpretation to display in
the next two sections.

\begin{figure*}[t]
\centering
\includegraphics[width=0.85\textwidth]{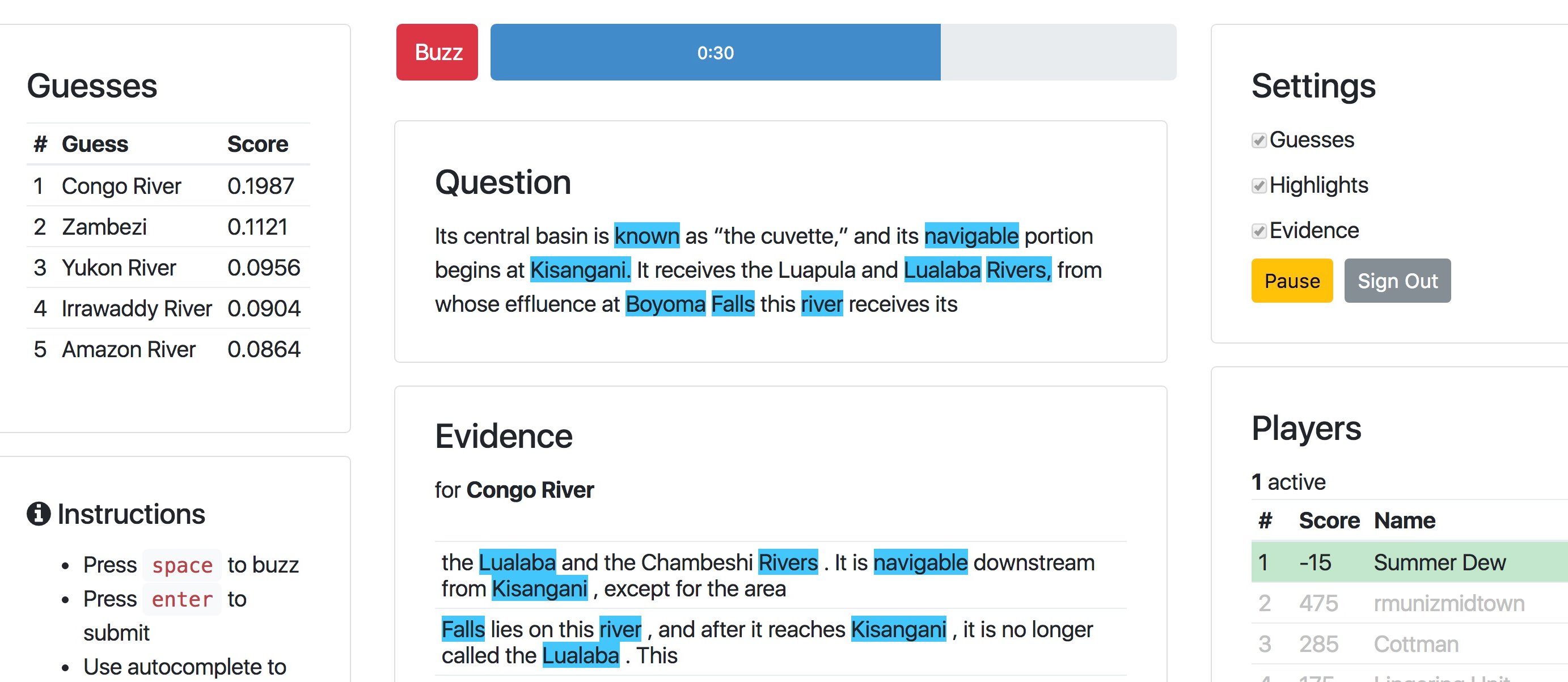}
\caption{\label{fig:screenshot} Screenshot of the interface. Question is
    displayed in the middle area word-by-word, with question highlights
    displayed in the same panel. Guesses are listed in the panel on the left.
    Evidence is in the panel below.}
\end{figure*}

\section{Interface Design}
\label{sec:design}

We design our \qb{} interface (Figure~\ref{fig:screenshot}) to
\emph{visualize} the three interpretations described in the previous
section. This section introduces the visualizations, placement, and
interactivity of the interface. 

To make \qb{} players feel at home, we follow the general framework
of~\url{Protobowl.com}, a popular \qb{} platform
that many
players actively use for practice.  The \textbf{Question} area is
in the center, and the question is displayed word-by-word in
the text box. A \textbf{Buzz} button is located close above the
question area, and to further reduce the distraction from the question
area, players can also buzz in using the space key. After buzzing, the
player have eight seconds to enter and select an answer from
a drop-down menu.

\begin{figure}[H]
\centering
\fbox{\includegraphics[width=0.4\columnwidth]{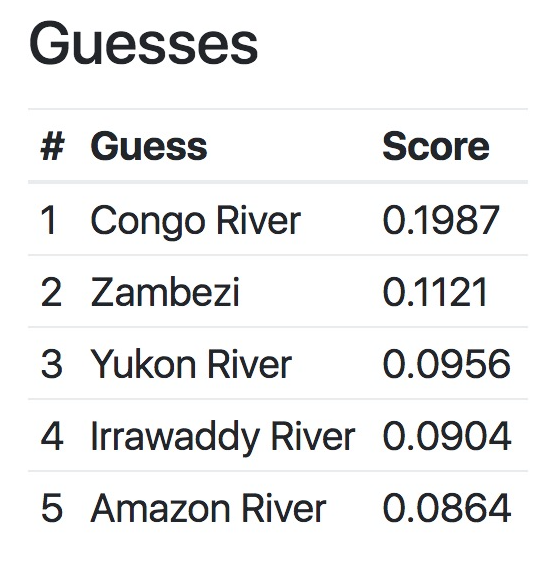}}
\end{figure}

\textbf{Guesses} show the answers the computer is considering along
with the associated score.  Top ten answers are sorted according to
their score (the system prefers higher scores).  This helps convey
when the model is uncertain (e.g., if all of the guesses have a low
score).

\begin{figure}[H]
\centering
\fbox{\includegraphics[width=0.85\columnwidth]{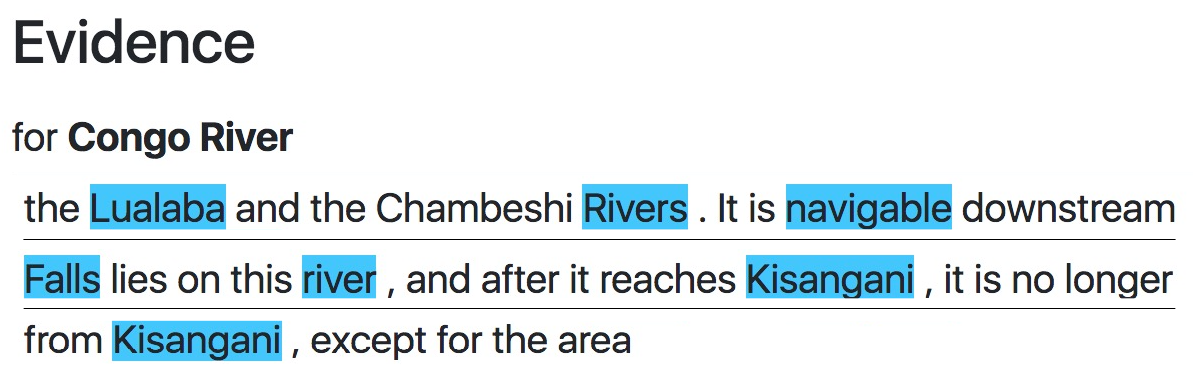}}
\end{figure}

To inform the player of how the model's prediction is supported by
training examples, \textbf{Evidence} shows the relevant snippets of
the most salient training examples for the top guess. It is located
below the question area and has the same width to provide a direct
comparison against the input question. Each line of the text area
shows the snippet of one selected document.

\begin{figure}[H]
\centering
\fbox{\includegraphics[width=0.85\columnwidth]{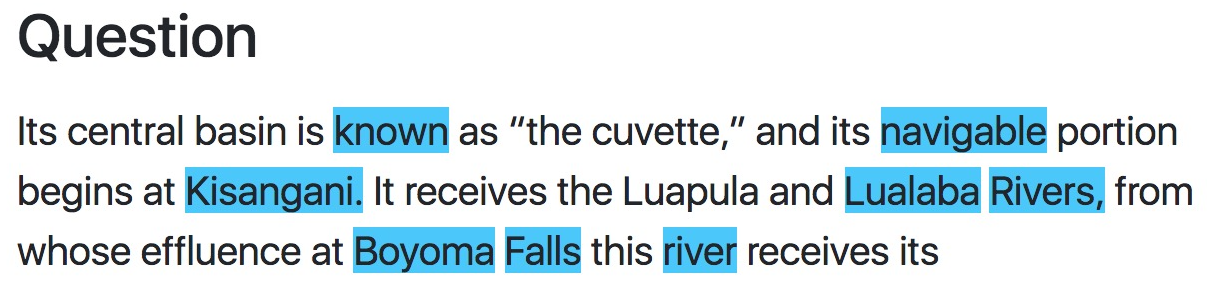}}
\end{figure}

We use \textbf{Highlight} to visualize the most salient words in both
the input question and the evidence snippets. These words are selected
for the top guess. As introduced in the previous section, we first
highlight important words in the training example snippets using an
\abr{api} of \abr{es}, then find their appearances in the input and
highlight those too. 


Multiple interpretations can be shown in combination. The
combination of highlight and evidence has a compounding effect: when
both are enabled, players see highlighted words in both the question
and the evidence (for example in Figure~\ref{fig:screenshot});
when highlight is enabled without evidence, players only see
highlights in the question.


Our design goal is to minimize distraction from the
question area while boosting the competitiveness of the player. So we place
the question area in the middle and have all interpretations around it.
It is difficult to ensure that different forms of
interpretations are exposed to the users equally, as some forms (e.g.,
evidence) are inherently less intuitive to visualize. However,
all interpretations must be implemented in an interface for a
real-world evaluation; we discuss the limitations of our design and
future work in Section~\ref{sec:discussion}.

\section{Setup}
\label{sec:setup}

This section explains how human players and the computer guesser play
in cooperation. To ensure accuracy and unbiasedness, we control what
interpretations each player sees instead of letting them choose.

\subsection{Data and Participants}

We collect 160 new questions for this evaluation that had not been 
previously seen by the \qb{} community to avoid bias in players'
exposure to questions.

We recruit 40 experts (\qb{} enthusiasts) by advertising on an
online forum, and 40 novices using MTurk. Experts are
free to play as many questions as they want (but each player can only
play a question once), and we encourage them to play more by offering
monetary prizes for those who finish the whole question set. We
require novices to each answer at least twenty questions and require
a positive score at the end (according to standard \qb{} scoring
rules) to encourage good faith responses. Online \qb{} platforms such
as \url{Protobowl.com} are usually anonymous, so we do not collect
any information about the participants other than an email
address for collecting prizes (optional).

\subsection{Human-AI Cooperation on \qb{}}

Unlike previous work where \qb{} interfaces are used for computers to
\emph{compete} with
humans~\cite{boydgraber2012besting,he2016opponent}, our interface aims
at human-\abr{ai} \emph{cooperation}. We let a human player form a
team with a computer teammate and put the human in charge.  As the
question is displayed word-by-word, the computer periodically updates
its guesses and interpretations (every 4 words in our experiments); at
any point before the question is fully read, the human can decide to
buzz, interrupt the readout, and provide an answer. The
interpretations should help the human better decide whether to
trust the computer's prediction or not.

We have two different experimental settings. In the simpler,
non-competitive \textbf{novice setting}, we have a single turker
interact with the interface, with the computer guesser
as teammate, but without opponents.

The competitive \textbf{expert setting} better resembles real \qb{}
games, and the players in this setting are experts
that enjoy the game.
To encourage them to play to the best of their ability, we
simulate the \qb{} setting as closely as possible (for novices the
simple task is already taxing enough without competition).  In a real
\qb{} match, players not just compete against themselves (can I get
the question right?) but also with each other (can I get the question
right before Selene does?).  \qb{}'s pyramidality
encourages competition: difficult clues at the start of the
question help determine who knows the most about a subject.
Our interface resembles \url{Protobowl.com}, a popular online \qb{}
platform where players play against each other (but
without the computer teammate).  The computer generates the same
output (both prediction and interpretations), but human players might
have access to different interpretations, e.g., David sees evidence while
Selene sees question highlights. Next section discusses the setup in
detail. 

Our experiment in the expert setting was possible thanks to \qb{}'s
enthusiast community. It was because \qb{}ers love to play this game
and to improve their skills by practicing, that they were willing to
learn our interface, team up with the computer, and compete under this
slightly irregular setting.  This provided us new perspectives of how
users from a wider range of skill levels use interpretations,
compared to many previous work that only had non-expert
turkers~\cite{smith2017evaluating,kneusel2017improving,clark2018creative}.

\subsection{Controlling Which Interpretations to Show}

Each of the three interpretations can be turned on or off, so we have
in total $2\times2\times2=8$ conditions, including the null condition
where all interpretations are hidden.
To compare within-subjects (players vary greatly based on their innate
ability), we vary the interpretations a player sees randomly.  We
sample the enabled combination with the goal of having, in
expectation, a uniform distribution over players, questions, and
interpretation combinations. For player $P$ at question $Q$, we sample
from an eight-class categorical distribution, with the parameter of
each combination~$C$ set to $N-\#(C, P)$, where $\#(C, P)$ is the
number of times player~$P$ has seen the interpretation combination~$C$
and $N$ is the expected count of each combination (in our case 
the number of questions divided by eight).  In the expert setting,
interpretations are sampled independently for each player, and players
may (and usually do) see different interpretations.
For all experiments, we only allow each player to answer each question
once.

\section{Results}
\label{sec:results}


\begin{figure}[t]
    \centering
    \textbf{Effect of interpretations}\par\medskip
    \includegraphics[width=1.0\columnwidth]{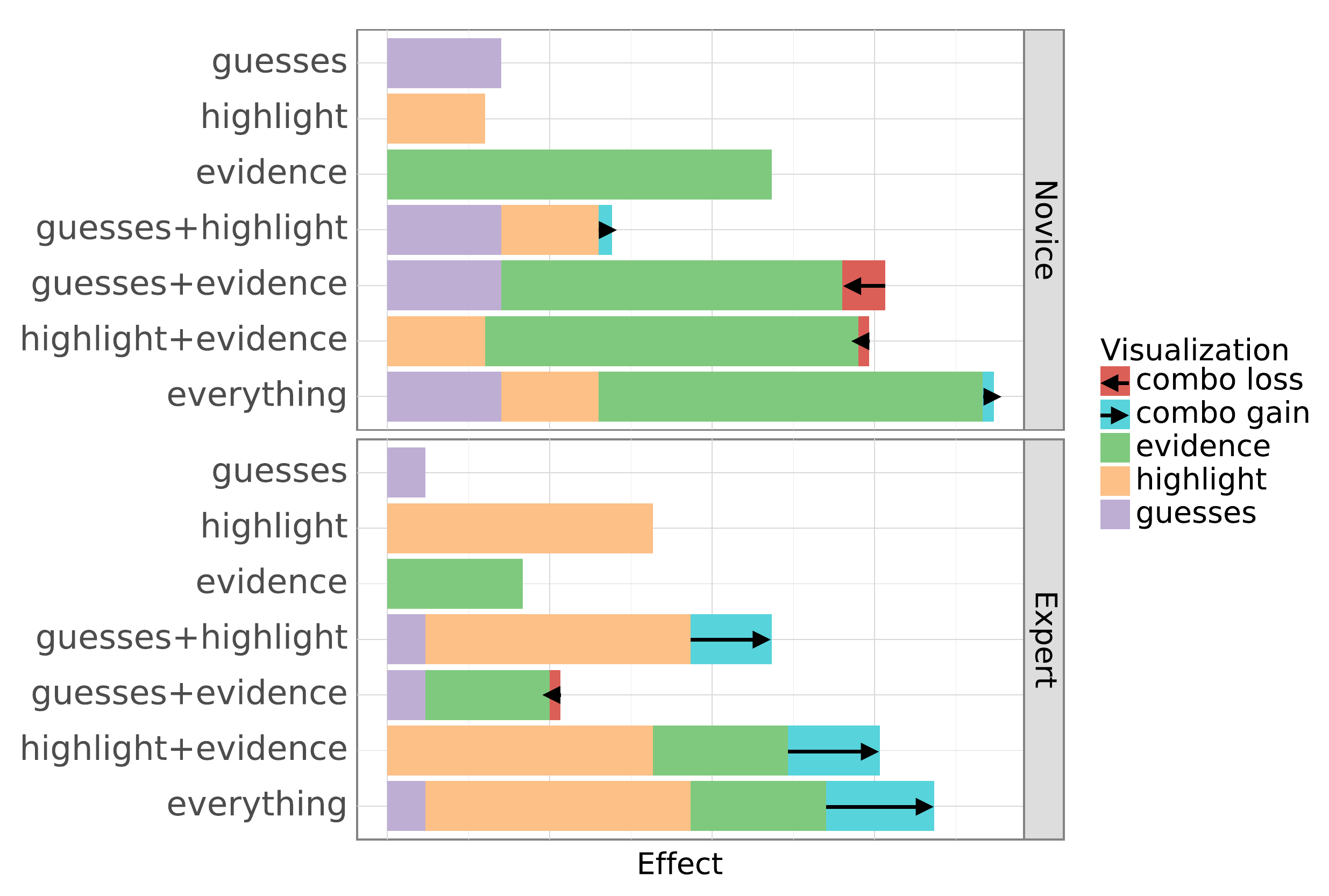}
    \caption{\label{fig:coefs} Coefficients of the linear regression
    showing the effects of interpretations, for novices (above) and
    experts (below).
    Higher value means an interpretation improves player accuracy.
    In addition to the individual interpretations, \emph{combo gain}
    and \emph{combo loss} capture the additional effect of combining
    multiple interpretations. \emph{Highlight} and \emph{Evidence}
    are effective for both novices and experts; combining leads to
    more positive effect for experts than novices, potentially because
    experts can process more information in limited time.}
\end{figure}

\begin{figure}[t]
    \centering
    \textbf{Effect of player ability}\par\medskip
    \includegraphics[width=.7\columnwidth]{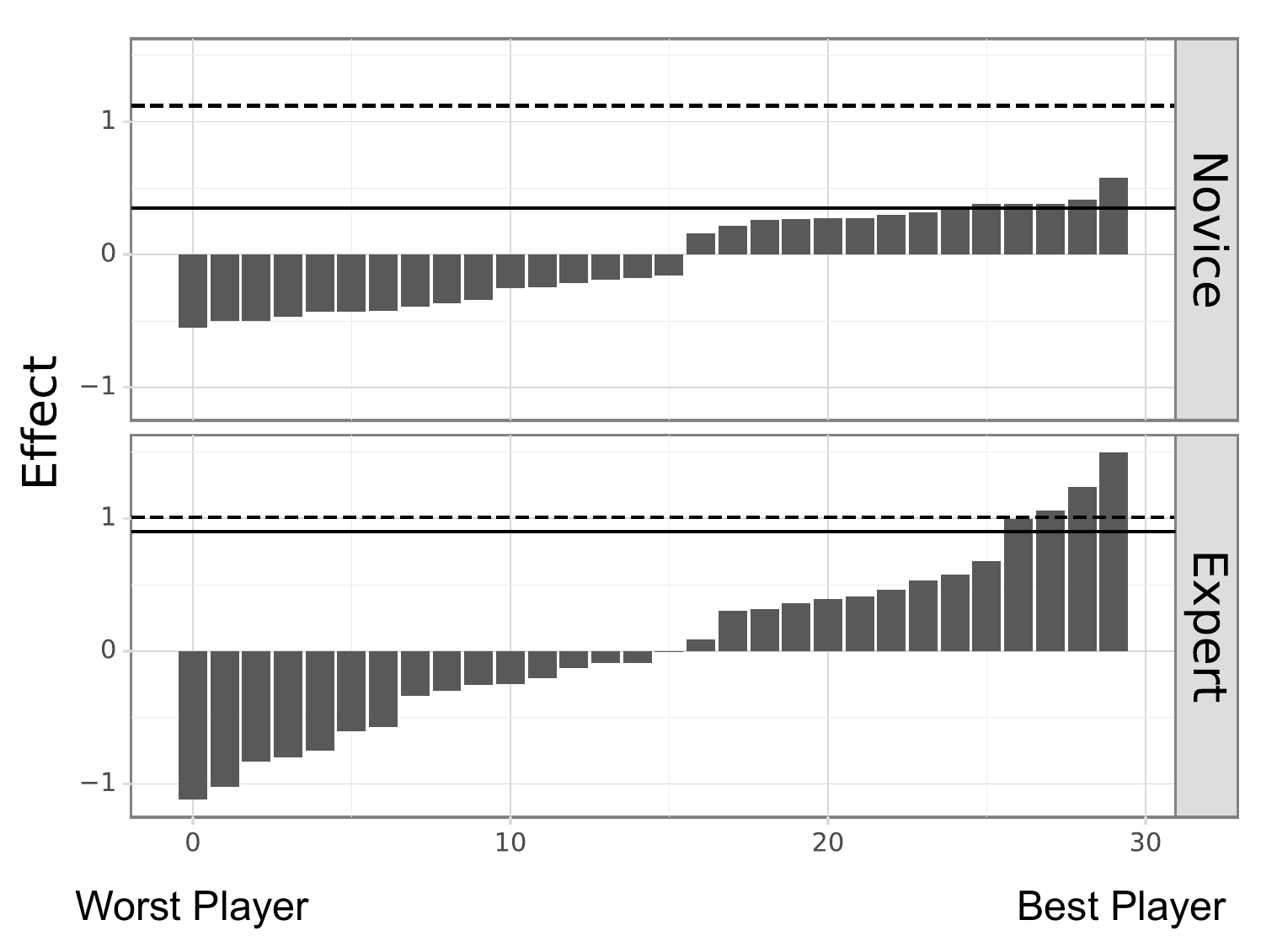}
    \par\medskip
    \textbf{Effect of question difficulty}\par\medskip
    \includegraphics[width=.7\columnwidth]{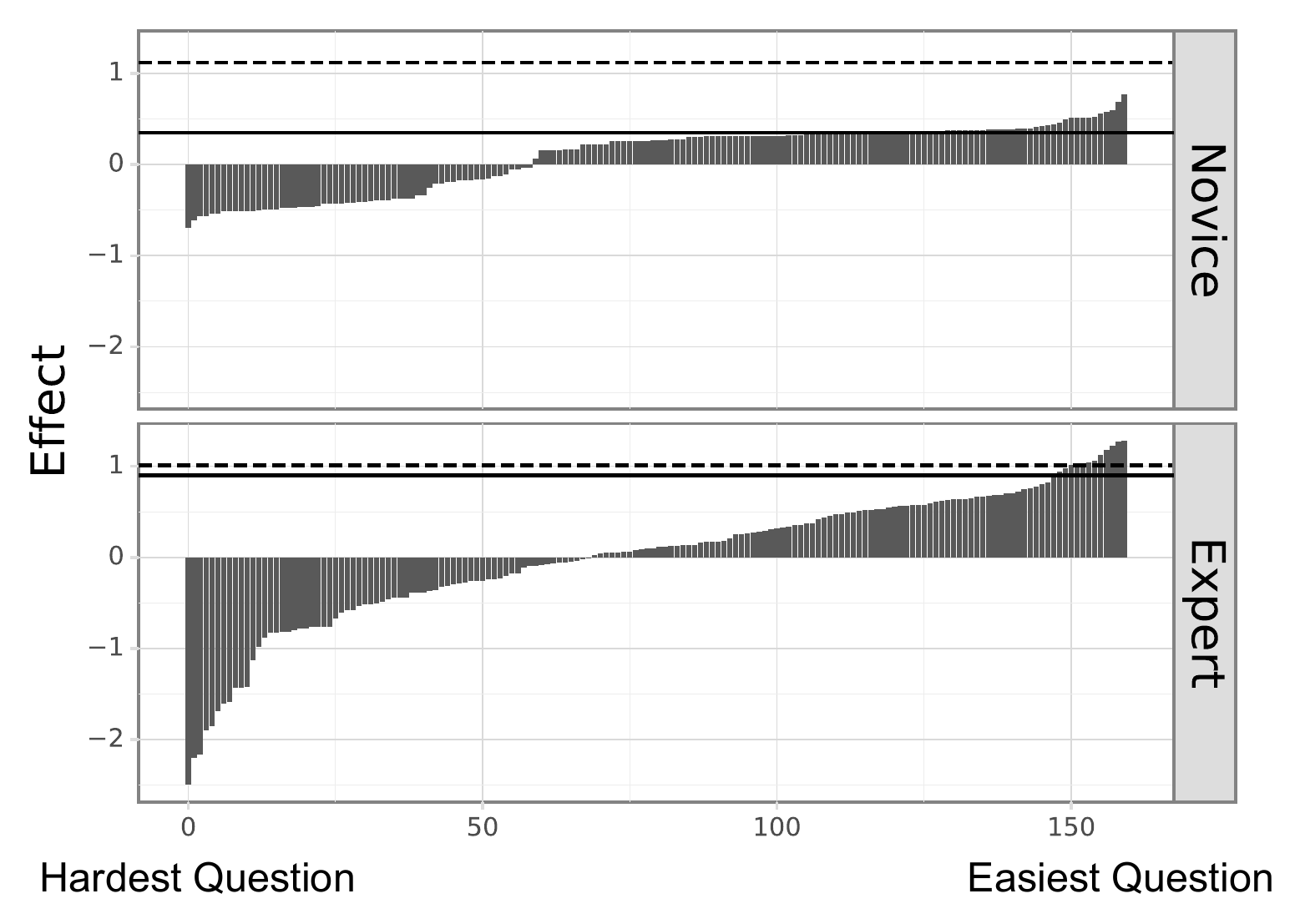}
    \caption{\label{fig:uq_coefs} Effect of player ability (above) and
    question difficulty (below) from the regression analysis. Solid
    horizontal lines show the bias term that captures the baseline
    accuracy without any help from the computer;
    dashed lines show the effect of combining all
    interpretations.  Experts have a higher average accuracy; they are
    also less affected by interpretations.}
\end{figure}


\begin{figure}[t]
    \centering
    \textbf{Distribution of buzzes}\par\medskip
    \includegraphics[width=\columnwidth]{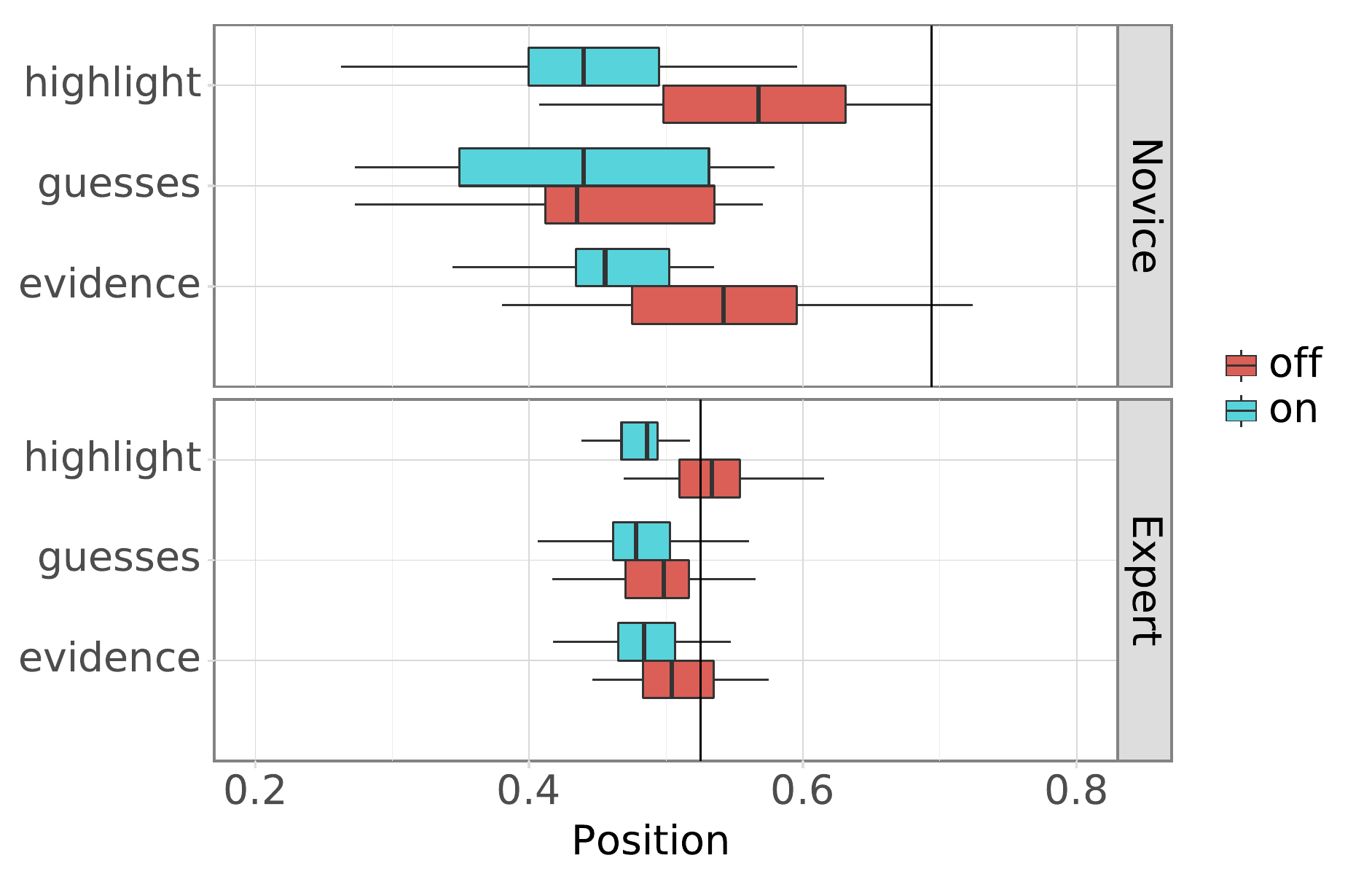}
    \caption{\label{fig:tools_buzz} Average buzzing position (relative
    to question length) of novices (above) and experts (below),
    with and without each interpretation. The goal is to buzz as early
    as possible.
    Vertical bars show the baseline buzzing position without any
    interpretation.
    Experts are better and more consistent.
    Among the interpretations, \emph{Highlight} is
    most effective in helping both novices and experts answer faster.}
\end{figure}

\begin{figure}[t]
\centering
\textbf{Aggressiveness of novice buzzes}\par\medskip
\includegraphics[width=.8\columnwidth]{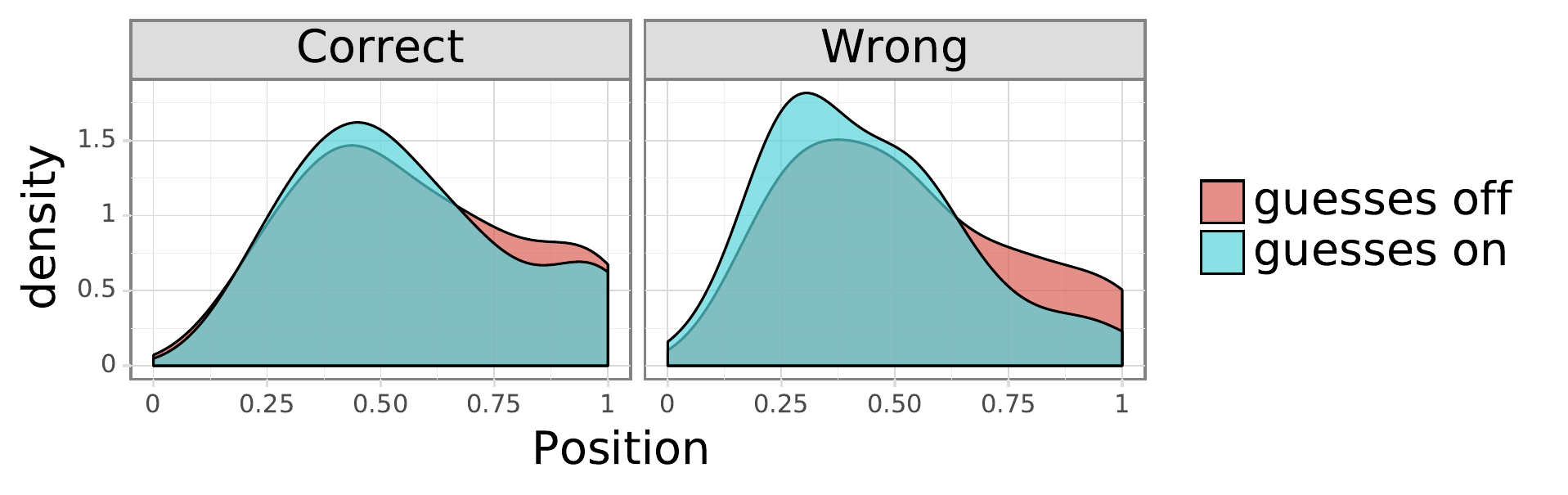}
\includegraphics[width=.8\columnwidth]{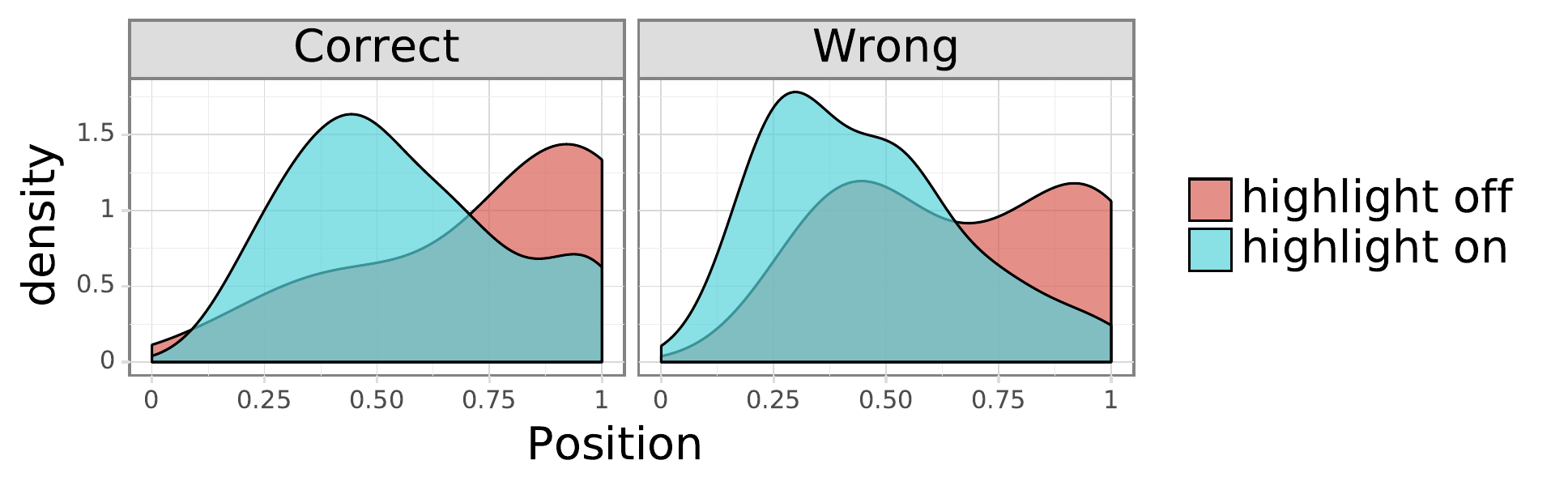}
\includegraphics[width=.8\columnwidth]{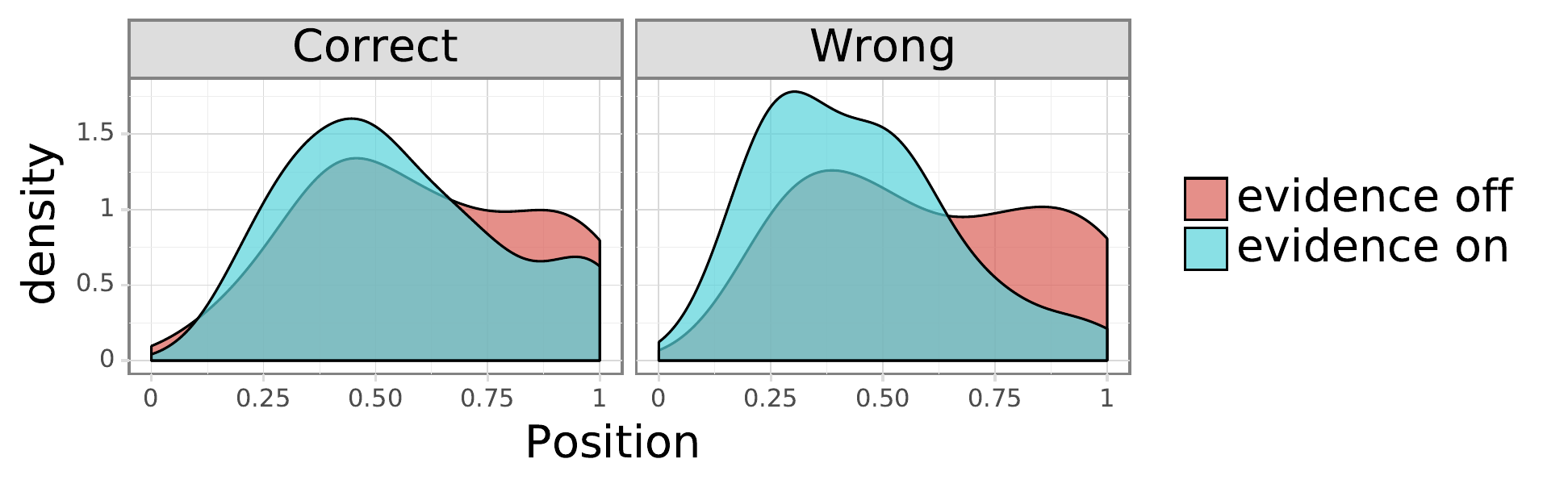}
\caption{\label{fig:aggressive_novice} The distribution of buzzes of
    novices on correct guesses (left) and wrong guesses (left); colors
    indicate if each interpretation is enabled; positions are
    normalized by question length.  With interpretations, novices are
    significantly more aggressive, but also get more questions correct
    earlier. \emph{Highlight} is the most effective.}
\end{figure}

\begin{figure}[t]
\centering
\textbf{Aggressiveness of expert buzzes}\par\medskip
\includegraphics[width=.8\columnwidth]{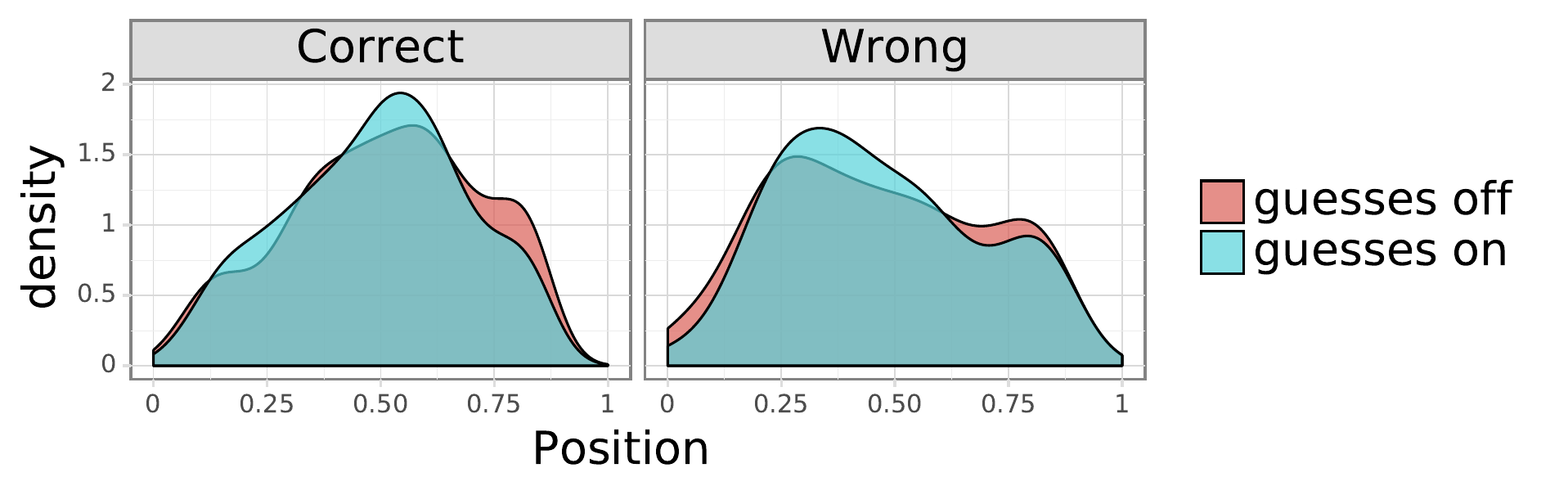}
\includegraphics[width=.8\columnwidth]{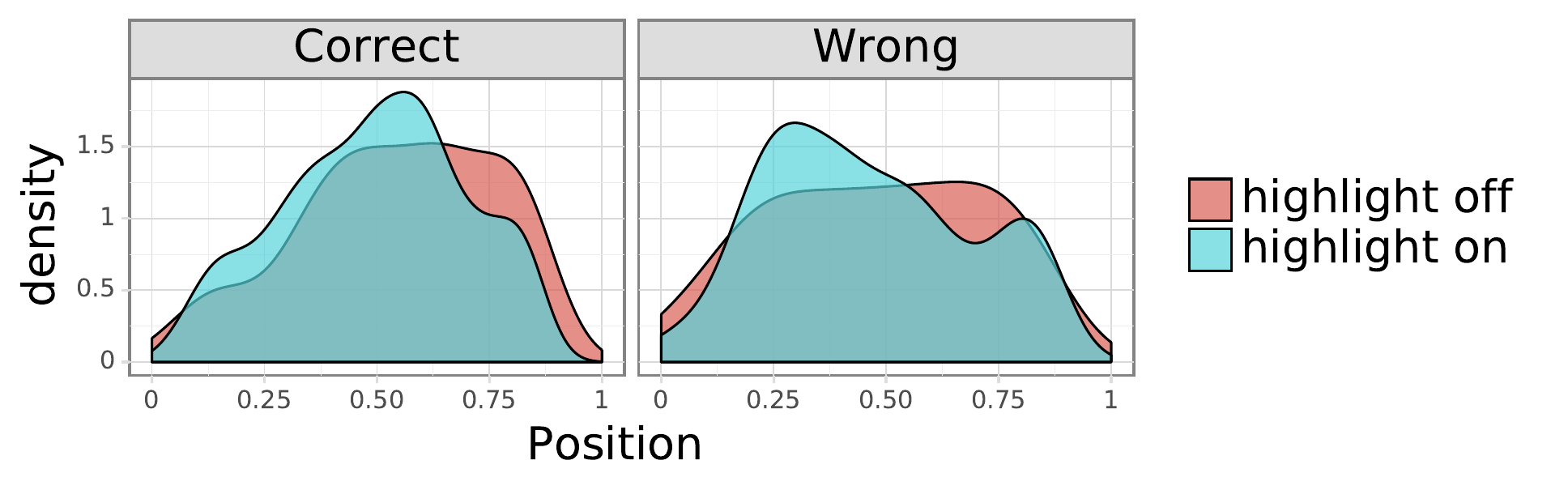}
\includegraphics[width=.8\columnwidth]{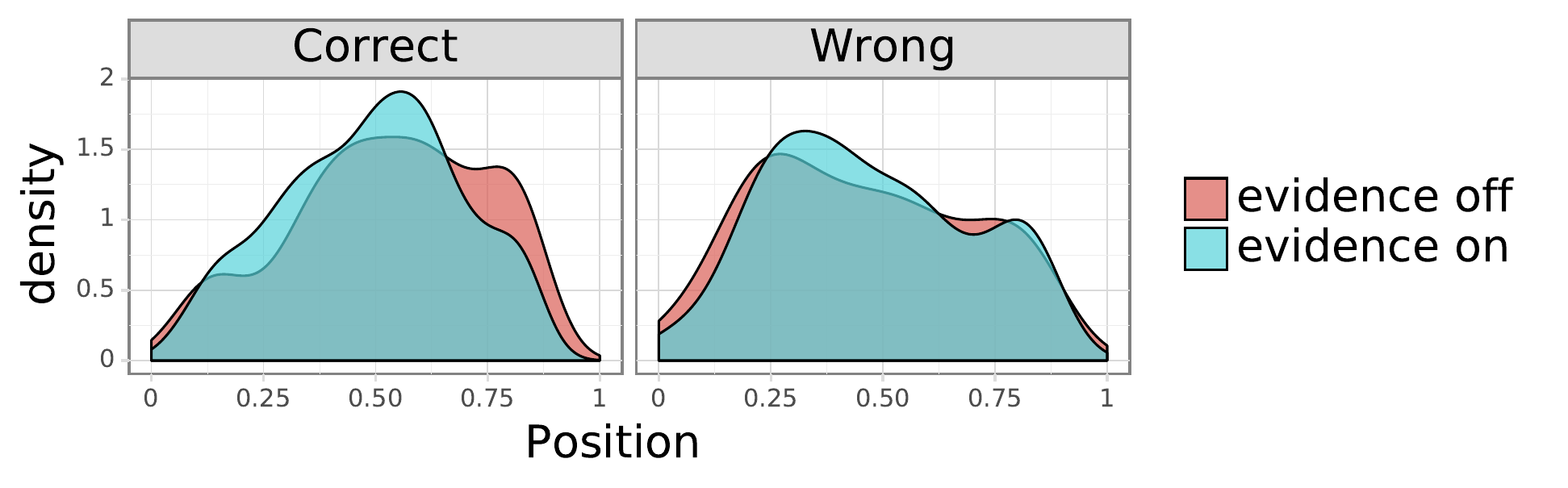}
    \caption{\label{fig:aggressive_expert} The distribution of buzzes
    of experts on correct guesses (left) and wrong guesses (left);
    colors indicate if each interpretation is enabled; positions are
    normalized by question length.  Experts are not significantly
    more aggressive with interpretations, but they did get more answers
    correct earlier.}
\end{figure}

With data collected from game plays, our primary goal is determine if
the interpretations are helpful or not, and how experts and novices
used them differently.
We first do a regression analysis to
quantitatively determine how much each condition affects the
accuracy of the players; then we break down the results to see
how the players behave differently under the conditions, specifically
how aggressive they are; we also look at specific cases where some
interpretation consistently succeeded or failed to convince multiple
players of the model prediction.

After filtering players who answer very few questions, we arrive at
30 experts that answer 1983 questions, and 30 novices that answer
600 questions. Turkers usually stopped after answering the required
twenty questions, but many experts kept on playing. Among all players,
seven experts answer all 160 questions.

\subsection{Regression Analysis}

Whether a player can answer a question correctly is determined by
several factors: the player's innate skill, the difficulty of the
question, the aid of some interpretation, or the competitive level (in
expert setting).  To tease apart these factors we follow
Narayanan~\etal{}~\cite{narayanan2018humans} and apply a regression
analysis.

\begin{table}
\begin{tabular}{l|l}
\multirow{5}{*}{\begin{tabular}[c]{@{}l@{}}Interpretation\\ (8)\end{tabular}} & none, guesses, highlight, evidence,     \\
                                                                              & guesses + highlight,                    \\
                                                                              & guesses + evidence,                     \\
                                                                              & highlight + evidence,                   \\
                                                                              & guesses + highlight + evidence          \\ \hline
\multirow{2}{*}{Player (30)}                                                  & player IDs                              \\
                                                                              & (separate for experts and novices)      \\ \hline
Question (160)                                                                & question IDs                            \\ \hline
\multirow{5}{*}{Others (3)}                                                   & buzzing position                        \\
                                                                              & (relative to question length),          \\
                                                                              & number of active players (expert only), \\
                                                                              & current accuracy of the top             \\
                                                                              & active player (expert only)
\end{tabular}
\caption{\label{table:features} Our four sets of features used in
regression analysis. Numbers in the parentheses indicate the number of
features in that set.}
\end{table}

We describe these factors using the four sets of features listed in
Table~\ref{table:features}.
To capture the player's innate skill and the difficulty of the
question, we include the IDs of both in the feature set.  Each
combination of interpretations has its own features, for example,
\emph{guesses}, \emph{evidence}, and \emph{guesses+evidence} are three
independent features.  For game condition, the first feature is the
relative position in the question when the player buzzed (to
understand how interpretations affect buzzing position as an outcome
instead of feature, we use a separate analysis); for the expert
setting, we also include extra features to capture the
competitiveness: number of active players and the current accuracy of
the top active player.

The we use a linear model to predict whether the player can answer the
question correctly.  Specifically, for each game record, we extract
the features and feed the vector as input to the linear model, which
then predicts the probability of a positive result; to train the
model, we compare the prediction against the ground truth, and update
the model with gradient descent.  We train this model on the game play
data, for experts and novices separately.

The coefficients of the linear model then explains the
importance of the corresponding features: the probability of a
positive result increases with features with positive coefficients, which
means these features help the players. Similarly negative coefficient
means the features hurt the player accuracy.  To understand which
interpretations are most helpful to \qb{} players, we inspect the sign
and magnitude of their corresponding coefficients.

Figure~\ref{fig:coefs} shows the effect of interpretations based on
regression coefficients: a high positive weight means the
interpretation is useful, zero means it is ineffective, and negative
means it is harmful.
It is not guaranteed that the strengths of multiple interpretations
are combined when they are displayed at the same time.
This is due to confounding factors such as information load---the
player might feel distracted when too much information is displayed on
the interface and thus perform worse.
The additional effects of combining
interpretations are ``combo gain'' and ``combo loss''
(Figure~\ref{fig:coefs}).
For example, combining guesses and evidence has a negative effect on
novices; the loss is computed by comparing the ``guesses+evidence''
coefficient with the arithmetic sum of the ``guesses'' and
``evidence'' coefficients.

The interpretation that helps novices is not the same as what helps
experts.  For experts, highlight is the most helpful individual
interpretation, while for novices, evidence is the most helpful. For
experts, the combination of highlight and evidence achieves extra
gain, which is reasonable because this combination adds
highlights to the evidence, making the contrast more intuitive.
However, the same combination does not show additional benefit for
novices, potentially due to information overload.

We hypothesize that the main difference between experts and novices is
that experts can use evidence more effectively. 
Question highlighting requires less
multitasking than evidence: players have to look away from the
question they need to answer to take in the evidence.  \qb{} players
likely know when they can glance down to related training data and can
also determine whether the training data are helpful.

To understand how much variance players display in their skill and
questions in their difficulty, we show their corresponding
coefficients (Figure~\ref{fig:uq_coefs}). The solid horizontal line
shows the baseline accuracy of that player group without any
interpretation (the bias term---or the intercept---of the linear
model). Experts show a higher baseline accuracy, which is not
surprising since they are experts; they also show a larger variance
in accuracy within the group, potentially due to the competitive
environment; they are also more sensitive to the difference in
question difficulty.
To compare these factors against the interpretations, we
show with the dashed horizontal line the combination of all three
interpretations. Experts are less sensitive to the interpretations,
potentially due to a higher confidence in their own guesses.

\subsection{How Interpretations Change Player Behavior}

The regression analysis provides a quantitative comparison between all
interpretations in how they affect the player accuracy. However,
accuracy alone does not tell the full story of how they play the
game. This section describes how each
interpretation affects the behavior of the players and how the effect
differs for novices and experts.  Ideal players should be both
aggressive and accurate: seeing very few words and answering
correctly. Interpretations should help them reach this goal.

Figure~\ref{fig:tools_buzz} show the
average buzzing position of each player group with and without each
interpretation.
Novices buzz much later than experts when no interpretation is enabled
(comparing the solid vertical bars), but buzz at about the same point
as experts when interpretations are enabled, despite a lower accuracy
(Figure~\ref{fig:uq_coefs}). This suggests that the novices are too
trusting in the computer teammate, and end up playing too aggressively
for their skill level.

We see a similar trend when we plot the density 
of buzzing positions (experts in Figure~\ref{fig:aggressive_expert}
and novices in Figure~\ref{fig:aggressive_novice}). In all settings,
the density shifts earlier: players are more aggressive with
interpretations, especially for novices, which is consistent with
Figure~\ref{fig:tools_buzz}. The interpretations allow players to
answer correctly earlier. Especially for novices with
highlights, the distribution of correct buzzing positions shifts
significantly earlier in the questions.

Although novices are helped by visualizations, these visualizations
are not enough to help them discern useful help from misleading help.
Novices are too aggressive at the start of the question with
visualizations: they trust the predictions of the system too much.
While experts mentally tune out bad suggestions, novices
are less discerning.  Visualizations thus 
must also convey whether they should be trusted, not just what answer
they are suggesting.

\subsection{Successes and Failures of Interpretations}

We now examine specific cases where interpretations help or hurt
players.

\begin{figure}[t]
\centering
\tikz\node[fill=white!70!colorsquad,inner sep=1pt,rounded corners=0.3cm]{
\begin{tabular}{p{0.46\textwidth}}
\emph{Question}:\\
(This essay) was composed after its author \textbf{refused} to
pay a \textbf{poll tax} to support the \textbf{Mexican-American}
\textbf{war}, and its ideology inspired Martin Luther King, Jr.\
and Mohandas Gandhi. \\

\emph{Evidence}:\\
him to pay six years of delinquent \textbf{poll tax}. Thoreau
\textbf{refused} because of his opposition to the
\textbf{Mexican-American War} and slavery, and he spent a night
in jail because of this refusal.
\end{tabular}
}; 
\caption{\label{fig:success} Interpretations that help players answer
a question on \underline{Civil Disobedience} correctly.  With the
shown part of the question, three experts answer correctly with the
evidence; no expert answer correctly without.}
\end{figure}

Figure~\ref{fig:success} shows an example where interpretations enable
players to answer correctly. A total of twelve expert players
answered the question, and eight answered correctly. The
earliest an expert can answer correctly without the evidence was at 72\%
of the question, while the three experts with the evidence all
answer correctly before 50\%. With the evidence and highlight,
players can infer from the keywords that the author is
Thoreau and that the guess is likely correct.  The computer shows
a salient training example and is effective in convincing the players
that the retrieved evidence is correct.

\begin{figure}[t]
\centering
\tikz\node[fill=white!70!colorsnli,inner sep=1pt,rounded corners=0.3cm]{
\begin{tabular}{p{0.46\textwidth}}
\emph{Question}:\\
A \textbf{book} by this man was first published with a
\textbf{preface} by Andreas Osiander titled \textbf{Ad Lectorem}.
\\
\emph{Evidence}:\\
the \textbf{Ad Lectorem} \textbf{preface} to Copernicus's
\textbf{book} was not actually by him.
\end{tabular}
}; 
\caption{\label{fig:fail_to_convince} Interpretations that fail to
    convince players.  Three expert players, when presented with the
    interpretation (some question text and evidence omitted), rejected the
    computer's correct guess (\underline{Copernicus}) and answered
    differently.}
\end{figure}

Figure~\ref{fig:fail_to_convince} shows a failure to convince,
where the combination of highlight and evidence fails to convince the
player of the computer's \emph{correct} guess: three expert players
rejected the computer's prediction and provided different answers,
relatively early in the question (before 50\%). The information
provided by the evidence is that Copernicus has a book with a
preface named Ad Lectorem, this piece of evidence strongly supports the
computer's guess \underline{Copernicus}.  However, it is expressed
differently than the question, with an unrelated but confusing ``not''
in the middle of the sentence.

%
%


\section{Discussion}
\label{sec:discussion}

%

%


The evaluation we present is grounded in a realistic setting, but also
task-specific. This section discusses how our method can be directly
applied to other settings, its limitations, and how we can incorporate
other components such as an eye tracker to our framework for a more
fine-grained assessment of interpretability.

\subsection{Forms and Methods of Interpretation}

Interpretations take on many forms, and within each form we have multiple
methods to generate the interpretation. For example, to highlight
salient input features for image classification, we can use variants
of input gradient~\cite{baehrens2010explain, simonyan2013deep}.
To optimize the generalizability of our results (despite being
task-specific) and demonstrate the flexibility of our method, we
focus on a comparison between forms of interpretation. To
select one method of each form, we choose a
high-performance linear model for its canonical interpretations.  Our
evaluation framework, including the interface and the regression
analysis, can be directly applied to a different comparison---one
between multiple methods of the same form.  This comparison is
particularly useful in the case of neural models, where all existing
interpretations are some approximation, and the evaluation of
how faithful they are to the model is crucial.

\subsection{Intrinsic and Extrinsic Evaluation}

Our approach is an extrinsic evaluation~\cite{narayanan2018humans}. The
task is played by thousands who compete in regularly.  Using
\qb{} allows a contextual, motivated evaluation of whether an
interpretation is useful. In contrast, intrinsic evaluation relies on
the interpretation alone. It is more direct but limited. In tasks
where no ground-truth explanation is available, the most tractable and
commonly used method is to construct ground-truth using a simpler
model as a benchmark for interpretability. For example, weights of linear models are
used for evaluating input highlight
explanations~\cite{li2016understanding,murdoch2018cd}.
This is restricted to tasks where the benchmark model performs
similarly to the complex model that requires interpretation, and it
does not work in application-grounded setting
(Section~\ref{sec:eval_qb}).

Extrinsic evaluations are hard to design, as they are affected by more
factors, especially humans' trust. When a user does not trust the
model and ignores it, the difference in the performance is not
affected by the explanations at all. Narayanan~\etal{}~\cite{narayanan2018humans} uses
``alien'' tasks to enforce trust, tasks that humans do not have
knowledge of. Our approach, in contrast, considers trust as an
inherent part of the cooperation: good interpretations should be
consistent and intuitive to convince humans to use it.


\subsection{Generalizing to Other Tasks}
Our method can be applied to natural language tasks other than
\qb{}, although \qb{}'s characters make it uniquely suitable.  To use
our interface for some other text classification task, for example
sentiment analysis or spam detection, one can convert the task into an
incremental version where the input is shown
word-by-word. Time limitation or competition can be added to
encourage the users to pay attention to
visualizations~\cite{narayanan2018humans}. One task
related to \qb{} has wide real world application: simultaneous
interpretation (or simultaneous translation, not to be
confused with model interpretation). Interpreters need to trade off
between accuracy and delay, much like \qb{}ers need to balance
accuracy and aggressiveness. The underlying mechanism of the
\abr{qanta} buzzer~\cite{he2016opponent} also resembles how
simultaneous translation systems handle this
trade-off~\cite{grissom2014don}.

\subsection{Limitations}
First, because we
compare visualizations individually and in combinations, their
placement is fixed to avoid confusing the players. The fixed
placement leads to uneven exposure to the users, so they might pay
less attention to some visualizations than others. If we focus on
individual visualizations, one way to resolve this issue is to display
the interpretation in a single fixed location, for example below the
question area. This would lead to a fair display of different
visualizations without confusing the users. However, one single
location might not suit all visualizations: for example, input
highlight should collocate with the input, while evidence is best
displayed next to the input for comparison.

Visualizations displayed on our interface change from question to
question, and the randomization (Setup)
might confuse the users. Before answering questions, each user sees
a tutorial that walks through the components of the interface,
but this can be improved by a set of warm-up questions to familiarize
the users of the interaction, which we will implement in future
studies. In addition, we can randomly sort the questions instead of
the visualizations, so the users see the same layout for multiple
questions, reducing context switches and consequently the cognitive
load.

Another limitation of our study is that, when a player's performance
improves with some interpretation, we cannot tell how much of that
improvement comes from the player using that interpretation.
We cannot derive causality from correlation. The key
missing factor is how much attention the player gave the interpretation,
and how much the decision is based on that. The attention the player
gave each interpretation could be measured using an eye tracker, and we
leave this to future work.

\subsection{Future Work}

While we focus on broad categories of interpretations to reveal that
some visualizations are more effective than others (e.g., highlighting
is more useful than guess lists), we can also use this approach to
evaluate specific highlighting methods in a task-based setting.  This
can help reveal how best to choose spans for highlighting, which words are best
suited for highlighting, and how to convey uncertainty in
highlighting.

While our evaluation focuses on the downstream task, we can expand
our analysis to measure how much users look at visualizations and in
what contexts (e.g., with an eye tracker).  This would reveal situational usefulness of
visualization components; if, for example, highlighting were only
useful to distinguish when two guesses had similar scores, we could
decrease cognitive load by only showing highlights when needed.

A tantalizing extension is to make these modifications automatically,
using the reward of task performance to encourage a reinforcement
learning algorithm to adjust interface elements to optimize
performance: such as changing font sizes, setting buttons for users to
explicitly agree or disagree with model predictions, or modifying the
highlighting strategy.

%
%

\section{Conclusion}
\label{sec:conclusion}

We propose and demonstrate an evaluation of interpretation methods in a
human-\abr{ai} cooperative setting. We focus on the natural language
domain and use a question answering task derived from a popular trivia
game, \qb{}.
Our experiments with both experts and novices reveal how they trust and
use interpretations differently, producing a more accurate and
realistic evaluation of machine learning interpretability. Our results
highlight the importance of taking the skill level of the target user
into consideration, and suggests that, combining
interpretations more intelligently and adapting to the user, we can
further improve the human-\abr{ai} cooperation.

\begin{acks}
We thank the anonymous reviewers for their insightful and constructive
comments.  Additionally, we would like to thank Alison Smith, Leah
Findlater, Hernisa Kacorri, Alina Striner, and Andreas Mathisen for
their valueble input.
This work was supported NSF Grant \abr{IIS}-1822494 and subcontract to
Raytheon BBN Technologies by DARPA award HR0011-15-C-0113.
Any opinions, findings,
conclusions, or recommendations expressed here are those of the
authors and do not necessarily reflect the view of the sponsor.

\end{acks}

\bibliographystyle{ACM-Reference-Format}
\bibliography{journal-full,main}


\begin{thebibliography}{68}


\ifx \showCODEN    \undefined \def \showCODEN     #1{\unskip}     \fi
\ifx \showDOI      \undefined \def \showDOI       #1{#1}\fi
\ifx \showISBNx    \undefined \def \showISBNx     #1{\unskip}     \fi
\ifx \showISBNxiii \undefined \def \showISBNxiii  #1{\unskip}     \fi
\ifx \showISSN     \undefined \def \showISSN      #1{\unskip}     \fi
\ifx \showLCCN     \undefined \def \showLCCN      #1{\unskip}     \fi
\ifx \shownote     \undefined \def \shownote      #1{#1}          \fi
\ifx \showarticletitle \undefined \def \showarticletitle #1{#1}   \fi
\ifx \showURL      \undefined \def \showURL       {\relax}        \fi
\providecommand\bibfield[2]{#2}
\providecommand\bibinfo[2]{#2}
\providecommand\natexlab[1]{#1}
\providecommand\showeprint[2][]{arXiv:#2}

\bibitem[\protect\citeauthoryear{Adebayo, Kim, Goodfellow, Gilmer, and
  Hardt}{Adebayo et~al\mbox{.}}{2018}]%
        {adebayo2018sanity}
\bibfield{author}{\bibinfo{person}{Julius Adebayo}, \bibinfo{person}{Been Kim},
  \bibinfo{person}{Ian Goodfellow}, \bibinfo{person}{Justin Gilmer}, {and}
  \bibinfo{person}{Moritz Hardt}.} \bibinfo{year}{2018}\natexlab{}.
\newblock \showarticletitle{Sanity Checks for Saliency Maps}. In
  \bibinfo{booktitle}{\emph{Proceedings of Advances in Neural Information
  Processing Systems}}.
\newblock


\bibitem[\protect\citeauthoryear{Allen, Guinn, and Horvtz}{Allen
  et~al\mbox{.}}{1999}]%
        {allen1999mixed}
\bibfield{author}{\bibinfo{person}{JE Allen}, \bibinfo{person}{Curry~I Guinn},
  {and} \bibinfo{person}{E Horvtz}.} \bibinfo{year}{1999}\natexlab{}.
\newblock \showarticletitle{Mixed-initiative interaction}.
\newblock \bibinfo{journal}{\emph{IEEE Intelligent Systems and their
  Applications}} (\bibinfo{year}{1999}).
\newblock


\bibitem[\protect\citeauthoryear{Antifakos, Kern, Schiele, and
  Schwaninger}{Antifakos et~al\mbox{.}}{2005}]%
        {antifakos2005towards}
\bibfield{author}{\bibinfo{person}{Stavros Antifakos}, \bibinfo{person}{Nicky
  Kern}, \bibinfo{person}{Bernt Schiele}, {and} \bibinfo{person}{Adrian
  Schwaninger}.} \bibinfo{year}{2005}\natexlab{}.
\newblock \showarticletitle{Towards improving trust in context-aware systems by
  displaying system confidence}. In \bibinfo{booktitle}{\emph{Proceedings of
  the international conference on Human-computer interaction with mobile
  devices and services}}.
\newblock


\bibitem[\protect\citeauthoryear{Baehrens, Schroeter, Harmeling, Kawanabe,
  Hansen, and M{\"u}ller}{Baehrens et~al\mbox{.}}{2010}]%
        {baehrens2010explain}
\bibfield{author}{\bibinfo{person}{David Baehrens}, \bibinfo{person}{Timon
  Schroeter}, \bibinfo{person}{Stefan Harmeling}, \bibinfo{person}{Motoaki
  Kawanabe}, \bibinfo{person}{Katja Hansen}, {and}
  \bibinfo{person}{Klaus-Robert M{\"u}ller}.} \bibinfo{year}{2010}\natexlab{}.
\newblock \showarticletitle{How to Explain Individual Classification
  Decisions}.
\newblock \bibinfo{journal}{\emph{Journal of Machine Learning Research}}
  (\bibinfo{year}{2010}).
\newblock


\bibitem[\protect\citeauthoryear{Boyd-Graber, Feng, and Rodriguez}{Boyd-Graber
  et~al\mbox{.}}{2018}]%
        {nips2018qbcomp}
\bibfield{author}{\bibinfo{person}{Jordan Boyd-Graber}, \bibinfo{person}{Shi
  Feng}, {and} \bibinfo{person}{Pedro Rodriguez}.}
  \bibinfo{year}{2018}\natexlab{}.
\newblock \showarticletitle{Human-Computer Question Answering: The Case for
  Quizbowl}.
\newblock \bibinfo{journal}{\emph{The NIPS '17 Competition: Building
  Intelligent Systems}} (\bibinfo{year}{2018}).
\newblock


\bibitem[\protect\citeauthoryear{Boyd-Graber, Satinoff, He, and {Daum{\'e}
  III}}{Boyd-Graber et~al\mbox{.}}{2012}]%
        {boydgraber2012besting}
\bibfield{author}{\bibinfo{person}{Jordan~L. Boyd-Graber},
  \bibinfo{person}{Brianna Satinoff}, \bibinfo{person}{He He}, {and}
  \bibinfo{person}{Hal {Daum{\'e} III}}.} \bibinfo{year}{2012}\natexlab{}.
\newblock \showarticletitle{Besting the Quiz Master: Crowdsourcing Incremental
  Classification Games}. In \bibinfo{booktitle}{\emph{Proceedings of Empirical
  Methods in Natural Language Processing}}.
\newblock


\bibitem[\protect\citeauthoryear{Chakraborti, Kambhampati, Scheutz, and
  Zhang}{Chakraborti et~al\mbox{.}}{2017}]%
        {chakraborti2017ai}
\bibfield{author}{\bibinfo{person}{Tathagata Chakraborti},
  \bibinfo{person}{Subbarao Kambhampati}, \bibinfo{person}{Matthias Scheutz},
  {and} \bibinfo{person}{Yu Zhang}.} \bibinfo{year}{2017}\natexlab{}.
\newblock \showarticletitle{{AI} challenges in human-robot cognitive teaming}.
\newblock \bibinfo{journal}{\emph{arXiv preprint arXiv:1707.04775}}
  (\bibinfo{year}{2017}).
\newblock


\bibitem[\protect\citeauthoryear{Clark, Ross, Tan, Ji, and Smith}{Clark
  et~al\mbox{.}}{2018}]%
        {clark2018creative}
\bibfield{author}{\bibinfo{person}{Elizabeth Clark},
  \bibinfo{person}{Anne~Spencer Ross}, \bibinfo{person}{Chenhao Tan},
  \bibinfo{person}{Yangfeng Ji}, {and} \bibinfo{person}{Noah~A Smith}.}
  \bibinfo{year}{2018}\natexlab{}.
\newblock \showarticletitle{Creative Writing with a Machine in the Loop: Case
  Studies on Slogans and Stories}. In \bibinfo{booktitle}{\emph{International
  Conference on Intelligent User Interfaces}}.
\newblock


\bibitem[\protect\citeauthoryear{Deng, Dong, Socher, Li, Li, and Fei-Fei}{Deng
  et~al\mbox{.}}{2009}]%
        {deng2009imagenet}
\bibfield{author}{\bibinfo{person}{Jia Deng}, \bibinfo{person}{Wei Dong},
  \bibinfo{person}{Richard Socher}, \bibinfo{person}{Li-Jia Li},
  \bibinfo{person}{Kai Li}, {and} \bibinfo{person}{Li Fei-Fei}.}
  \bibinfo{year}{2009}\natexlab{}.
\newblock \showarticletitle{{ImageNet}: A large-scale hierarchical image
  database}. In \bibinfo{booktitle}{\emph{Computer Vision and Pattern
  Recognition}}.
\newblock


\bibitem[\protect\citeauthoryear{Doshi-Velez and Kim}{Doshi-Velez and
  Kim}{2018}]%
        {doshivelez2017towards}
\bibfield{author}{\bibinfo{person}{Finale Doshi-Velez} {and}
  \bibinfo{person}{Been Kim}.} \bibinfo{year}{2018}\natexlab{}.
\newblock \showarticletitle{Towards A Rigorous Science of Interpretable Machine
  Learning}.
\newblock \bibinfo{journal}{\emph{Springer Series on Challenges in Machine
  Learning}} (\bibinfo{year}{2018}).
\newblock


\bibitem[\protect\citeauthoryear{{European Parliament and Council of the
  European Union}}{{European Parliament and Council of the European
  Union}}{2016}]%
        {gdpr}
\bibfield{author}{\bibinfo{person}{{European Parliament and Council of the
  European Union}}.} \bibinfo{year}{2016}\natexlab{}.
\newblock \showarticletitle{General data protection regulation}.
\newblock  (\bibinfo{year}{2016}).
\newblock


\bibitem[\protect\citeauthoryear{Feng, Wallace, {Grissom II}, Iyyer, Rodriguez,
  and Boyd-Graber}{Feng et~al\mbox{.}}{2018}]%
        {feng2018rawr}
\bibfield{author}{\bibinfo{person}{Shi Feng}, \bibinfo{person}{Eric Wallace},
  \bibinfo{person}{Alvin {Grissom II}}, \bibinfo{person}{Mohit Iyyer},
  \bibinfo{person}{Pedro Rodriguez}, {and} \bibinfo{person}{Jordan
  Boyd-Graber}.} \bibinfo{year}{2018}\natexlab{}.
\newblock \showarticletitle{Pathologies of Neural Models Make Interpretations
  Difficult}. In \bibinfo{booktitle}{\emph{Proceedings of Empirical Methods in
  Natural Language Processing}}.
\newblock


\bibitem[\protect\citeauthoryear{Fong and Vedaldi}{Fong and Vedaldi}{2017}]%
        {fong2017interpretable}
\bibfield{author}{\bibinfo{person}{Ruth~C Fong} {and} \bibinfo{person}{Andrea
  Vedaldi}.} \bibinfo{year}{2017}\natexlab{}.
\newblock \showarticletitle{Interpretable explanations of black boxes by
  meaningful perturbation}. In \bibinfo{booktitle}{\emph{International
  Conference on Computer Vision}}.
\newblock


\bibitem[\protect\citeauthoryear{Ghorbani, Abid, and Zou}{Ghorbani
  et~al\mbox{.}}{2018}]%
        {ghorbani2017interpretation}
\bibfield{author}{\bibinfo{person}{Amirata Ghorbani}, \bibinfo{person}{Abubakar
  Abid}, {and} \bibinfo{person}{James~Y. Zou}.}
  \bibinfo{year}{2018}\natexlab{}.
\newblock \showarticletitle{Interpretation of Neural Networks is Fragile}. In
  \bibinfo{booktitle}{\emph{Association for the Advancement of Artificial
  Intelligence}}.
\newblock


\bibitem[\protect\citeauthoryear{Goodfellow, Shlens, and Szegedy}{Goodfellow
  et~al\mbox{.}}{2015}]%
        {goodfellow2014explaining}
\bibfield{author}{\bibinfo{person}{Ian~J. Goodfellow},
  \bibinfo{person}{Jonathon Shlens}, {and} \bibinfo{person}{Christian
  Szegedy}.} \bibinfo{year}{2015}\natexlab{}.
\newblock \showarticletitle{Explaining and Harnessing Adversarial Examples}. In
  \bibinfo{booktitle}{\emph{Proceedings of the International Conference on
  Learning Representations}}.
\newblock


\bibitem[\protect\citeauthoryear{Gormley and Tong}{Gormley and Tong}{2015}]%
        {gormley2015es}
\bibfield{author}{\bibinfo{person}{Clinton Gormley} {and}
  \bibinfo{person}{Zachary Tong}.} \bibinfo{year}{2015}\natexlab{}.
\newblock \bibinfo{booktitle}{\emph{Elasticsearch: The Definitive Guide}}.
\newblock \bibinfo{publisher}{O'Reilly Media, Inc.}
\newblock


\bibitem[\protect\citeauthoryear{Grissom~II, He, Boyd-Graber, Morgan, and
  Daum{\'e}~III}{Grissom~II et~al\mbox{.}}{2014}]%
        {grissom2014don}
\bibfield{author}{\bibinfo{person}{Alvin Grissom~II}, \bibinfo{person}{He He},
  \bibinfo{person}{Jordan Boyd-Graber}, \bibinfo{person}{John Morgan}, {and}
  \bibinfo{person}{Hal Daum{\'e}~III}.} \bibinfo{year}{2014}\natexlab{}.
\newblock \showarticletitle{Don’t until the final verb wait: Reinforcement
  learning for simultaneous machine translation}. In
  \bibinfo{booktitle}{\emph{Proceedings of Empirical Methods in Natural
  Language Processing}}.
\newblock


\bibitem[\protect\citeauthoryear{Grossman, Fitzmaurice, and Attar}{Grossman
  et~al\mbox{.}}{2009}]%
        {grossman2009survey}
\bibfield{author}{\bibinfo{person}{Tovi Grossman}, \bibinfo{person}{George
  Fitzmaurice}, {and} \bibinfo{person}{Ramtin Attar}.}
  \bibinfo{year}{2009}\natexlab{}.
\newblock \showarticletitle{A survey of software learnability: metrics,
  methodologies and guidelines}. In \bibinfo{booktitle}{\emph{International
  Conference on Human Factors in Computing Systems}}.
\newblock


\bibitem[\protect\citeauthoryear{Gunning}{Gunning}{2017}]%
        {gunning2017explainable}
\bibfield{author}{\bibinfo{person}{David Gunning}.}
  \bibinfo{year}{2017}\natexlab{}.
\newblock \showarticletitle{Explainable artificial intelligence ({XAI})}.
\newblock \bibinfo{journal}{\emph{Defense Advanced Research Projects Agency
  (DARPA), nd Web}} (\bibinfo{year}{2017}).
\newblock


\bibitem[\protect\citeauthoryear{Guo, Pleiss, Sun, and Weinberger}{Guo
  et~al\mbox{.}}{2017}]%
        {guo2017calibration}
\bibfield{author}{\bibinfo{person}{Chuan Guo}, \bibinfo{person}{Geoff Pleiss},
  \bibinfo{person}{Yu Sun}, {and} \bibinfo{person}{Kilian~Q. Weinberger}.}
  \bibinfo{year}{2017}\natexlab{}.
\newblock \showarticletitle{On Calibration of Modern Neural Networks}. In
  \bibinfo{booktitle}{\emph{Proceedings of the International Conference of
  Machine Learning}}.
\newblock


\bibitem[\protect\citeauthoryear{He, Boyd-Graber, Kwok, and {Daum{\'e} III}}{He
  et~al\mbox{.}}{2016}]%
        {he2016opponent}
\bibfield{author}{\bibinfo{person}{He He}, \bibinfo{person}{Jordan~L.
  Boyd-Graber}, \bibinfo{person}{Kevin Kwok}, {and} \bibinfo{person}{Hal
  {Daum{\'e} III}}.} \bibinfo{year}{2016}\natexlab{}.
\newblock \showarticletitle{Opponent Modeling in Deep Reinforcement Learning}.
  In \bibinfo{booktitle}{\emph{Proceedings of the International Conference of
  Machine Learning}}.
\newblock


\bibitem[\protect\citeauthoryear{He, Zhang, Ren, and Sun}{He
  et~al\mbox{.}}{2015}]%
        {he2015delving}
\bibfield{author}{\bibinfo{person}{Kaiming He}, \bibinfo{person}{Xiangyu
  Zhang}, \bibinfo{person}{Shaoqing Ren}, {and} \bibinfo{person}{Jian Sun}.}
  \bibinfo{year}{2015}\natexlab{}.
\newblock \showarticletitle{Delving deep into rectifiers: Surpassing
  human-level performance on imagenet classification}. In
  \bibinfo{booktitle}{\emph{International Conference on Computer Vision}}.
\newblock


\bibitem[\protect\citeauthoryear{Hooker, Erhan, Kindermans, and Kim}{Hooker
  et~al\mbox{.}}{2018}]%
        {hooker2018evaluating}
\bibfield{author}{\bibinfo{person}{Sara Hooker}, \bibinfo{person}{Dumitru
  Erhan}, \bibinfo{person}{Pieter-Jan Kindermans}, {and} \bibinfo{person}{Been
  Kim}.} \bibinfo{year}{2018}\natexlab{}.
\newblock \showarticletitle{Evaluating Feature Importance Estimates}. In
  \bibinfo{booktitle}{\emph{ICML Workshop on Human Interpretability in Machine
  Learning}}.
\newblock


\bibitem[\protect\citeauthoryear{Horvitz}{Horvitz}{1999}]%
        {horvitz1999principles}
\bibfield{author}{\bibinfo{person}{Eric Horvitz}.}
  \bibinfo{year}{1999}\natexlab{}.
\newblock \showarticletitle{Principles of mixed-initiative user interfaces}. In
  \bibinfo{booktitle}{\emph{International Conference on Human Factors in
  Computing Systems}}.
\newblock


\bibitem[\protect\citeauthoryear{Iyyer, Boyd-Graber, Claudino, Socher, and
  {Daum{\'e} III}}{Iyyer et~al\mbox{.}}{2014}]%
        {iyyer2014ann}
\bibfield{author}{\bibinfo{person}{Mohit Iyyer}, \bibinfo{person}{Jordan
  Boyd-Graber}, \bibinfo{person}{Leonardo Max~Batista Claudino},
  \bibinfo{person}{Richard Socher}, {and} \bibinfo{person}{Hal {Daum{\'e}
  III}}.} \bibinfo{year}{2014}\natexlab{}.
\newblock \showarticletitle{A Neural Network for Factoid Question Answering
  over Paragraphs}. In \bibinfo{booktitle}{\emph{Proceedings of Empirical
  Methods in Natural Language Processing}}.
\newblock


\bibitem[\protect\citeauthoryear{Jia and Liang}{Jia and Liang}{2017}]%
        {jia2017adversarial}
\bibfield{author}{\bibinfo{person}{Robin Jia} {and} \bibinfo{person}{Percy
  Liang}.} \bibinfo{year}{2017}\natexlab{}.
\newblock \showarticletitle{Adversarial Examples for Evaluating Reading
  Comprehension Systems}. In \bibinfo{booktitle}{\emph{Proceedings of Empirical
  Methods in Natural Language Processing}}.
\newblock


\bibitem[\protect\citeauthoryear{Jiang, Kim, and Gupta}{Jiang
  et~al\mbox{.}}{2018}]%
        {jiang2018trust}
\bibfield{author}{\bibinfo{person}{Heinrich Jiang}, \bibinfo{person}{Been Kim},
  {and} \bibinfo{person}{Maya~R. Gupta}.} \bibinfo{year}{2018}\natexlab{}.
\newblock \showarticletitle{To Trust Or Not To Trust A Classifier}. In
  \bibinfo{booktitle}{\emph{Proceedings of Advances in Neural Information
  Processing Systems}}.
\newblock


\bibitem[\protect\citeauthoryear{Ju and Leifer}{Ju and Leifer}{2008}]%
        {ju2008design}
\bibfield{author}{\bibinfo{person}{Wendy Ju} {and} \bibinfo{person}{Larry
  Leifer}.} \bibinfo{year}{2008}\natexlab{}.
\newblock \showarticletitle{The design of implicit interactions: Making
  interactive systems less obnoxious}.
\newblock \bibinfo{journal}{\emph{Design Issues}} (\bibinfo{year}{2008}).
\newblock


\bibitem[\protect\citeauthoryear{Kindermans, Hooker, Adebayo, Alber,
  Sch{\"u}tt, D{\"a}hne, Erhan, and Kim}{Kindermans et~al\mbox{.}}{2017}]%
        {kindermans2017unreliability}
\bibfield{author}{\bibinfo{person}{Pieter-Jan Kindermans},
  \bibinfo{person}{Sara Hooker}, \bibinfo{person}{Julius Adebayo},
  \bibinfo{person}{Maximilian Alber}, \bibinfo{person}{Kristof~T. Sch{\"u}tt},
  \bibinfo{person}{Sven D{\"a}hne}, \bibinfo{person}{Dumitru Erhan}, {and}
  \bibinfo{person}{Been Kim}.} \bibinfo{year}{2017}\natexlab{}.
\newblock \showarticletitle{The (Un)reliability of saliency methods}.
\newblock \bibinfo{journal}{\emph{arXiv preprint arXiv: 1711.00867}}
  (\bibinfo{year}{2017}).
\newblock


\bibitem[\protect\citeauthoryear{Kleinberg, Lakkaraju, Leskovec, Ludwig, and
  Mullainathan}{Kleinberg et~al\mbox{.}}{2017}]%
        {kleinberg2017human}
\bibfield{author}{\bibinfo{person}{Jon Kleinberg}, \bibinfo{person}{Himabindu
  Lakkaraju}, \bibinfo{person}{Jure Leskovec}, \bibinfo{person}{Jens Ludwig},
  {and} \bibinfo{person}{Sendhil Mullainathan}.}
  \bibinfo{year}{2017}\natexlab{}.
\newblock \showarticletitle{Human decisions and machine predictions}.
\newblock \bibinfo{journal}{\emph{The quarterly journal of economics}}
  (\bibinfo{year}{2017}).
\newblock


\bibitem[\protect\citeauthoryear{Kneusel and Mozer}{Kneusel and Mozer}{2017}]%
        {kneusel2017improving}
\bibfield{author}{\bibinfo{person}{Ronald~T Kneusel} {and}
  \bibinfo{person}{Michael~C Mozer}.} \bibinfo{year}{2017}\natexlab{}.
\newblock \showarticletitle{Improving Human-Machine Cooperative Visual Search
  With Soft Highlighting}.
\newblock \bibinfo{journal}{\emph{ACM Transactions on Applied Perception}}
  (\bibinfo{year}{2017}).
\newblock


\bibitem[\protect\citeauthoryear{Koedinger, Brunskill, Baker, McLaughlin, and
  Stamper}{Koedinger et~al\mbox{.}}{2013}]%
        {Koedinger-13}
\bibfield{author}{\bibinfo{person}{Kenneth~R. Koedinger}, \bibinfo{person}{Emma
  Brunskill}, \bibinfo{person}{Ryan~S.J.d. Baker},
  \bibinfo{person}{Elizabeth~A. McLaughlin}, {and} \bibinfo{person}{John
  Stamper}.} \bibinfo{year}{2013}\natexlab{}.
\newblock \showarticletitle{New Potentials for Data-Driven Intelligent Tutoring
  System Development and Optimization}.
\newblock \bibinfo{journal}{\emph{{AI} Magazine}} \bibinfo{volume}{34},
  \bibinfo{number}{3} (\bibinfo{date}{sep} \bibinfo{year}{2013}),
  \bibinfo{pages}{27}.
\newblock
\urldef\tempurl%
\url{https://doi.org/10.1609/aimag.v34i3.2484}
\showDOI{\tempurl}


\bibitem[\protect\citeauthoryear{Koh and Liang}{Koh and Liang}{2017}]%
        {koh2017influence}
\bibfield{author}{\bibinfo{person}{Pang~Wei Koh} {and} \bibinfo{person}{Percy
  Liang}.} \bibinfo{year}{2017}\natexlab{}.
\newblock \showarticletitle{Understanding Black-box Predictions via Influence
  Functions}. In \bibinfo{booktitle}{\emph{Proceedings of the International
  Conference of Machine Learning}}.
\newblock


\bibitem[\protect\citeauthoryear{Krause, Perer, and Ng}{Krause
  et~al\mbox{.}}{2016}]%
        {krause2016interacting}
\bibfield{author}{\bibinfo{person}{Josua Krause}, \bibinfo{person}{Adam Perer},
  {and} \bibinfo{person}{Kenney Ng}.} \bibinfo{year}{2016}\natexlab{}.
\newblock \showarticletitle{Interacting with predictions: Visual inspection of
  black-box machine learning models}. In
  \bibinfo{booktitle}{\emph{International Conference on Human Factors in
  Computing Systems}}.
\newblock


\bibitem[\protect\citeauthoryear{Lai and Tan}{Lai and Tan}{2019}]%
        {lai2018human}
\bibfield{author}{\bibinfo{person}{Vivian Lai} {and} \bibinfo{person}{Chenhao
  Tan}.} \bibinfo{year}{2019}\natexlab{}.
\newblock \showarticletitle{On Human Predictions with Explanations and
  Predictions of Machine Learning Models: A Case Study on Deception Detection}.
  In \bibinfo{booktitle}{\emph{Proceedings of ACM FAT*}}.
\newblock


\bibitem[\protect\citeauthoryear{Lakkaraju, Bach, and Leskovec}{Lakkaraju
  et~al\mbox{.}}{2016}]%
        {lakkaraju2016interpretable}
\bibfield{author}{\bibinfo{person}{Himabindu Lakkaraju},
  \bibinfo{person}{Stephen~H Bach}, {and} \bibinfo{person}{Jure Leskovec}.}
  \bibinfo{year}{2016}\natexlab{}.
\newblock \showarticletitle{Interpretable decision sets: A joint framework for
  description and prediction}. In \bibinfo{booktitle}{\emph{Knowledge Discovery
  and Data Mining}}.
\newblock


\bibitem[\protect\citeauthoryear{LeCun, Bengio, and Hinton}{LeCun
  et~al\mbox{.}}{2015}]%
        {lecun2015deep}
\bibfield{author}{\bibinfo{person}{Yann LeCun}, \bibinfo{person}{Yoshua
  Bengio}, {and} \bibinfo{person}{Geoffrey Hinton}.}
  \bibinfo{year}{2015}\natexlab{}.
\newblock \showarticletitle{Deep learning}.
\newblock \bibinfo{journal}{\emph{Nature}} (\bibinfo{year}{2015}).
\newblock


\bibitem[\protect\citeauthoryear{Lee, Smith, Seppi, Elmqvist, Boyd-Graber, and
  Findlater}{Lee et~al\mbox{.}}{2017}]%
        {lee2017human}
\bibfield{author}{\bibinfo{person}{Tak~Yeon Lee}, \bibinfo{person}{Alison
  Smith}, \bibinfo{person}{Kevin Seppi}, \bibinfo{person}{Niklas Elmqvist},
  \bibinfo{person}{Jordan Boyd-Graber}, {and} \bibinfo{person}{Leah
  Findlater}.} \bibinfo{year}{2017}\natexlab{}.
\newblock \showarticletitle{The human touch: How non-expert users perceive,
  interpret, and fix topic models}.
\newblock \bibinfo{journal}{\emph{International Journal of Human-Computer
  Studies}} (\bibinfo{year}{2017}).
\newblock


\bibitem[\protect\citeauthoryear{Letham, Rudin, McCormick, Madigan,
  et~al\mbox{.}}{Letham et~al\mbox{.}}{2015}]%
        {letham2015interpretable}
\bibfield{author}{\bibinfo{person}{Benjamin Letham}, \bibinfo{person}{Cynthia
  Rudin}, \bibinfo{person}{Tyler~H McCormick}, \bibinfo{person}{David Madigan},
  {et~al\mbox{.}}} \bibinfo{year}{2015}\natexlab{}.
\newblock \showarticletitle{Interpretable classifiers using rules and bayesian
  analysis: Building a better stroke prediction model}.
\newblock \bibinfo{journal}{\emph{The Annals of Applied Statistics}}
  (\bibinfo{year}{2015}).
\newblock


\bibitem[\protect\citeauthoryear{Li, Monroe, and Jurafsky}{Li
  et~al\mbox{.}}{2016}]%
        {li2016understanding}
\bibfield{author}{\bibinfo{person}{Jiwei Li}, \bibinfo{person}{Will Monroe},
  {and} \bibinfo{person}{Daniel Jurafsky}.} \bibinfo{year}{2016}\natexlab{}.
\newblock \showarticletitle{Understanding Neural Networks through
  Representation Erasure}.
\newblock \bibinfo{journal}{\emph{arXiv preprint arXiv: 1612.08220}}
  (\bibinfo{year}{2016}).
\newblock


\bibitem[\protect\citeauthoryear{Lipton}{Lipton}{2016}]%
        {lipton2016mythos}
\bibfield{author}{\bibinfo{person}{Zachary~Chase Lipton}.}
  \bibinfo{year}{2016}\natexlab{}.
\newblock \showarticletitle{The Mythos of Model Interpretability}.
\newblock \bibinfo{journal}{\emph{arXiv preprint arXiv: 1606.03490}}
  (\bibinfo{year}{2016}).
\newblock


\bibitem[\protect\citeauthoryear{Liu, Wang, Liu, and Zhu}{Liu
  et~al\mbox{.}}{2017}]%
        {liu2017towards}
\bibfield{author}{\bibinfo{person}{Shixia Liu}, \bibinfo{person}{Xiting Wang},
  \bibinfo{person}{Mengchen Liu}, {and} \bibinfo{person}{Jun Zhu}.}
  \bibinfo{year}{2017}\natexlab{}.
\newblock \showarticletitle{Towards better analysis of machine learning models:
  A visual analytics perspective}.
\newblock \bibinfo{journal}{\emph{Visual Informatics}} (\bibinfo{year}{2017}).
\newblock


\bibitem[\protect\citeauthoryear{Marcus, Marcinkiewicz, and Santorini}{Marcus
  et~al\mbox{.}}{1993}]%
        {marcus1993building}
\bibfield{author}{\bibinfo{person}{Mitchell~P Marcus},
  \bibinfo{person}{Mary~Ann Marcinkiewicz}, {and} \bibinfo{person}{Beatrice
  Santorini}.} \bibinfo{year}{1993}\natexlab{}.
\newblock \showarticletitle{Building a large annotated corpus of English: The
  Penn Treebank}.
\newblock \bibinfo{journal}{\emph{Computational linguistics}}
  (\bibinfo{year}{1993}).
\newblock


\bibitem[\protect\citeauthoryear{Miller}{Miller}{2017}]%
        {miller2017explanation}
\bibfield{author}{\bibinfo{person}{Tim Miller}.}
  \bibinfo{year}{2017}\natexlab{}.
\newblock \showarticletitle{Explanation in artificial intelligence: insights
  from the social sciences}.
\newblock \bibinfo{journal}{\emph{arXiv preprint arXiv:1706.07269}}
  (\bibinfo{year}{2017}).
\newblock


\bibitem[\protect\citeauthoryear{Mnih, Kavukcuoglu, Silver, Rusu, Veness,
  Bellemare, Graves, Riedmiller, Fidjeland, Ostrovski, et~al\mbox{.}}{Mnih
  et~al\mbox{.}}{2015}]%
        {mnih2015human}
\bibfield{author}{\bibinfo{person}{Volodymyr Mnih}, \bibinfo{person}{Koray
  Kavukcuoglu}, \bibinfo{person}{David Silver}, \bibinfo{person}{Andrei~A
  Rusu}, \bibinfo{person}{Joel Veness}, \bibinfo{person}{Marc~G Bellemare},
  \bibinfo{person}{Alex Graves}, \bibinfo{person}{Martin Riedmiller},
  \bibinfo{person}{Andreas~K Fidjeland}, \bibinfo{person}{Georg Ostrovski},
  {et~al\mbox{.}}} \bibinfo{year}{2015}\natexlab{}.
\newblock \showarticletitle{Human-level control through deep reinforcement
  learning}.
\newblock \bibinfo{journal}{\emph{Nature}} (\bibinfo{year}{2015}).
\newblock


\bibitem[\protect\citeauthoryear{Murdoch, Liu, and Yu}{Murdoch
  et~al\mbox{.}}{2018}]%
        {murdoch2018cd}
\bibfield{author}{\bibinfo{person}{W.~James Murdoch}, \bibinfo{person}{Peter~J.
  Liu}, {and} \bibinfo{person}{Bin Yu}.} \bibinfo{year}{2018}\natexlab{}.
\newblock \showarticletitle{Beyond Word Importance: Contextual Decomposition to
  Extract Interactions from LSTMs}. In \bibinfo{booktitle}{\emph{Proceedings of
  the International Conference on Learning Representations}}.
\newblock


\bibitem[\protect\citeauthoryear{Narayanan, Chen, He, Kim, Gershman, and
  Doshi-Velez}{Narayanan et~al\mbox{.}}{2018}]%
        {narayanan2018humans}
\bibfield{author}{\bibinfo{person}{Menaka Narayanan}, \bibinfo{person}{Emily
  Chen}, \bibinfo{person}{Jeffrey He}, \bibinfo{person}{Been Kim},
  \bibinfo{person}{Sam Gershman}, {and} \bibinfo{person}{Finale Doshi-Velez}.}
  \bibinfo{year}{2018}\natexlab{}.
\newblock \showarticletitle{How do Humans Understand Explanations from Machine
  Learning Systems? An Evaluation of the Human-Interpretability of
  Explanation}.
\newblock \bibinfo{journal}{\emph{arXiv preprint arXiv: 1802.00682}}
  (\bibinfo{year}{2018}).
\newblock


\bibitem[\protect\citeauthoryear{Papandrea, Raganato, and Bovi}{Papandrea
  et~al\mbox{.}}{2017}]%
        {papandrea2017supwsd}
\bibfield{author}{\bibinfo{person}{Simone Papandrea},
  \bibinfo{person}{Alessandro Raganato}, {and} \bibinfo{person}{Claudio~Delli
  Bovi}.} \bibinfo{year}{2017}\natexlab{}.
\newblock \showarticletitle{SUPWSD: A Flexible Toolkit for Supervised Word
  Sense Disambiguation}. In \bibinfo{booktitle}{\emph{Proceedings of the
  Conference on Empirical Methods in Natural Language Processing: System
  Demonstrations}}.
\newblock


\bibitem[\protect\citeauthoryear{Papernot and McDaniel}{Papernot and
  McDaniel}{2018}]%
        {papernot2018dknn}
\bibfield{author}{\bibinfo{person}{Nicolas Papernot} {and}
  \bibinfo{person}{Patrick~D. McDaniel}.} \bibinfo{year}{2018}\natexlab{}.
\newblock \showarticletitle{Deep k-Nearest Neighbors: Towards Confident,
  Interpretable and Robust Deep Learning}.
\newblock \bibinfo{journal}{\emph{arXiv preprint arXiv: 1803.04765}}
  (\bibinfo{year}{2018}).
\newblock


\bibitem[\protect\citeauthoryear{Peters, V{\"a}stfj{\"a}ll, Slovic, Mertz,
  Mazzocco, and Dickert}{Peters et~al\mbox{.}}{2006}]%
        {peters2006numeracy}
\bibfield{author}{\bibinfo{person}{Ellen Peters}, \bibinfo{person}{Daniel
  V{\"a}stfj{\"a}ll}, \bibinfo{person}{Paul Slovic}, \bibinfo{person}{CK
  Mertz}, \bibinfo{person}{Ketti Mazzocco}, {and} \bibinfo{person}{Stephan
  Dickert}.} \bibinfo{year}{2006}\natexlab{}.
\newblock \showarticletitle{Numeracy and decision making}.
\newblock \bibinfo{journal}{\emph{Psychological science}}
  (\bibinfo{year}{2006}).
\newblock


\bibitem[\protect\citeauthoryear{Reyna and Brainerd}{Reyna and
  Brainerd}{2008}]%
        {reyna2008numeracy}
\bibfield{author}{\bibinfo{person}{Valerie~F Reyna} {and}
  \bibinfo{person}{Charles~J Brainerd}.} \bibinfo{year}{2008}\natexlab{}.
\newblock \showarticletitle{Numeracy, ratio bias, and denominator neglect in
  judgments of risk and probability}.
\newblock \bibinfo{journal}{\emph{Learning and individual differences}}
  (\bibinfo{year}{2008}).
\newblock


\bibitem[\protect\citeauthoryear{Ribeiro, Singh, and Guestrin}{Ribeiro
  et~al\mbox{.}}{2016}]%
        {ribeiro2016lime}
\bibfield{author}{\bibinfo{person}{Marco~T{\'u}lio Ribeiro},
  \bibinfo{person}{Sameer Singh}, {and} \bibinfo{person}{Carlos Guestrin}.}
  \bibinfo{year}{2016}\natexlab{}.
\newblock \showarticletitle{Why Should I Trust You?": Explaining the
  Predictions of Any Classifier}. In \bibinfo{booktitle}{\emph{Knowledge
  Discovery and Data Mining}}.
\newblock


\bibitem[\protect\citeauthoryear{Ribeiro, Singh, and Guestrin}{Ribeiro
  et~al\mbox{.}}{2018}]%
        {ribeiro2018semantically}
\bibfield{author}{\bibinfo{person}{Marco~Tulio Ribeiro},
  \bibinfo{person}{Sameer Singh}, {and} \bibinfo{person}{Carlos Guestrin}.}
  \bibinfo{year}{2018}\natexlab{}.
\newblock \showarticletitle{Semantically Equivalent Adversarial Rules for
  Debugging NLP Models}. In \bibinfo{booktitle}{\emph{Proceedings of the
  Association for Computational Linguistics}}.
\newblock


\bibitem[\protect\citeauthoryear{Ross and Doshi-Velez}{Ross and
  Doshi-Velez}{2018}]%
        {ross2018regularizing}
\bibfield{author}{\bibinfo{person}{Andrew~Slavin Ross} {and}
  \bibinfo{person}{Finale Doshi-Velez}.} \bibinfo{year}{2018}\natexlab{}.
\newblock \showarticletitle{Improving the Adversarial Robustness and
  Interpretability of Deep Neural Networks by Regularizing their Input
  Gradients}. In \bibinfo{booktitle}{\emph{Association for the Advancement of
  Artificial Intelligence}}.
\newblock


\bibitem[\protect\citeauthoryear{Ross, Gordon, and Bagnell}{Ross
  et~al\mbox{.}}{2011}]%
        {ross2011reduction}
\bibfield{author}{\bibinfo{person}{St{\'e}phane Ross},
  \bibinfo{person}{Geoffrey Gordon}, {and} \bibinfo{person}{Drew Bagnell}.}
  \bibinfo{year}{2011}\natexlab{}.
\newblock \showarticletitle{A reduction of imitation learning and structured
  prediction to no-regret online learning}. In
  \bibinfo{booktitle}{\emph{Proceedings of Artificial Intelligence and
  Statistics}}.
\newblock


\bibitem[\protect\citeauthoryear{Rudin}{Rudin}{2018}]%
        {rudin2018please}
\bibfield{author}{\bibinfo{person}{Cynthia Rudin}.}
  \bibinfo{year}{2018}\natexlab{}.
\newblock \showarticletitle{Please Stop Explaining Black Box Models for High
  Stakes Decisions}. In \bibinfo{booktitle}{\emph{NIPS 2018 Workshop on
  Critiquing and Correcting Trends in Machine Learning}}.
\newblock


\bibitem[\protect\citeauthoryear{Rukzio, Hamard, Noda, and De~Luca}{Rukzio
  et~al\mbox{.}}{2006}]%
        {rukzio2006visualization}
\bibfield{author}{\bibinfo{person}{Enrico Rukzio}, \bibinfo{person}{John
  Hamard}, \bibinfo{person}{Chie Noda}, {and} \bibinfo{person}{Alexander
  De~Luca}.} \bibinfo{year}{2006}\natexlab{}.
\newblock \showarticletitle{Visualization of uncertainty in context aware
  mobile applications}. In \bibinfo{booktitle}{\emph{Proceedings of the
  international conference on Human-computer interaction with mobile devices
  and services}}.
\newblock


\bibitem[\protect\citeauthoryear{Schmidt and Herrmann}{Schmidt and
  Herrmann}{2017}]%
        {schmidt2017intervention}
\bibfield{author}{\bibinfo{person}{Albrecht Schmidt} {and}
  \bibinfo{person}{Thomas Herrmann}.} \bibinfo{year}{2017}\natexlab{}.
\newblock \showarticletitle{Intervention user interfaces: a new interaction
  paradigm for automated systems}.
\newblock \bibinfo{journal}{\emph{Interactions}} (\bibinfo{year}{2017}).
\newblock


\bibitem[\protect\citeauthoryear{Silver, Schrittwieser, Simonyan, Antonoglou,
  Huang, Guez, Hubert, Baker, Lai, Bolton, et~al\mbox{.}}{Silver
  et~al\mbox{.}}{2017}]%
        {silver2017mastering}
\bibfield{author}{\bibinfo{person}{David Silver}, \bibinfo{person}{Julian
  Schrittwieser}, \bibinfo{person}{Karen Simonyan}, \bibinfo{person}{Ioannis
  Antonoglou}, \bibinfo{person}{Aja Huang}, \bibinfo{person}{Arthur Guez},
  \bibinfo{person}{Thomas Hubert}, \bibinfo{person}{Lucas Baker},
  \bibinfo{person}{Matthew Lai}, \bibinfo{person}{Adrian Bolton},
  {et~al\mbox{.}}} \bibinfo{year}{2017}\natexlab{}.
\newblock \showarticletitle{Mastering the game of Go without human knowledge}.
\newblock \bibinfo{journal}{\emph{Nature}} (\bibinfo{year}{2017}).
\newblock


\bibitem[\protect\citeauthoryear{Simonyan, Vedaldi, and Zisserman}{Simonyan
  et~al\mbox{.}}{2014}]%
        {simonyan2013deep}
\bibfield{author}{\bibinfo{person}{Karen Simonyan}, \bibinfo{person}{Andrea
  Vedaldi}, {and} \bibinfo{person}{Andrew Zisserman}.}
  \bibinfo{year}{2014}\natexlab{}.
\newblock \showarticletitle{Deep Inside Convolutional Networks: Visualising
  Image Classification Models and Saliency Maps}. In
  \bibinfo{booktitle}{\emph{Proceedings of the International Conference on
  Learning Representations}}.
\newblock


\bibitem[\protect\citeauthoryear{Smith, Lee, Poursabzi-Sangdeh, Boyd-Graber,
  Elmqvist, and Findlater}{Smith et~al\mbox{.}}{2017}]%
        {smith2017evaluating}
\bibfield{author}{\bibinfo{person}{Alison Smith}, \bibinfo{person}{Tak~Yeon
  Lee}, \bibinfo{person}{Forough Poursabzi-Sangdeh}, \bibinfo{person}{Jordan
  Boyd-Graber}, \bibinfo{person}{Niklas Elmqvist}, {and} \bibinfo{person}{Leah
  Findlater}.} \bibinfo{year}{2017}\natexlab{}.
\newblock \showarticletitle{Evaluating visual representations for topic
  understanding and their effects on manually generated labels}.
\newblock \bibinfo{journal}{\emph{Transactions of the Association for
  Computational Linguistics}} (\bibinfo{year}{2017}).
\newblock


\bibitem[\protect\citeauthoryear{Sundararajan, Taly, and Yan}{Sundararajan
  et~al\mbox{.}}{2017}]%
        {sundararajan2017axiomatic}
\bibfield{author}{\bibinfo{person}{Mukund Sundararajan}, \bibinfo{person}{Ankur
  Taly}, {and} \bibinfo{person}{Qiqi Yan}.} \bibinfo{year}{2017}\natexlab{}.
\newblock \showarticletitle{Axiomatic Attribution for Deep Networks}. In
  \bibinfo{booktitle}{\emph{Proceedings of the International Conference of
  Machine Learning}}.
\newblock


\bibitem[\protect\citeauthoryear{Sutton and Barto}{Sutton and Barto}{1998}]%
        {sutton1998introduction}
\bibfield{author}{\bibinfo{person}{Richard~S Sutton} {and}
  \bibinfo{person}{Andrew~G Barto}.} \bibinfo{year}{1998}\natexlab{}.
\newblock \bibinfo{booktitle}{\emph{Introduction to reinforcement learning}}.
\newblock


\bibitem[\protect\citeauthoryear{Swartout}{Swartout}{1983}]%
        {swartout1983xplain}
\bibfield{author}{\bibinfo{person}{William~R Swartout}.}
  \bibinfo{year}{1983}\natexlab{}.
\newblock \bibinfo{booktitle}{\emph{Xplain: A system for creating and
  explaining expert consulting programs.}}
\newblock \bibinfo{type}{{T}echnical {R}eport}.
  \bibinfo{institution}{University of Southern California}.
\newblock


\bibitem[\protect\citeauthoryear{Thompson}{Thompson}{2013}]%
        {Thompson-13}
\bibfield{author}{\bibinfo{person}{Clive Thompson}.}
  \bibinfo{year}{2013}\natexlab{}.
\newblock \bibinfo{booktitle}{\emph{Smarter Than You Think: How Technology is
  Changing Our Minds for the Better}}.
\newblock \bibinfo{publisher}{The Penguin Group}.
\newblock
\showISBNx{1594204454, 9781594204456}


\bibitem[\protect\citeauthoryear{USACM}{USACM}{2017}]%
        {acm2017public}
\bibfield{author}{\bibinfo{person}{USACM}.} \bibinfo{year}{2017}\natexlab{}.
\newblock \showarticletitle{Statement on algorithmic transparency and
  accountability}.
\newblock \bibinfo{journal}{\emph{Public Policy Council}}
  (\bibinfo{year}{2017}).
\newblock


\bibitem[\protect\citeauthoryear{Vinson, Takayama, Forlizzi, Ju, Cakmak, and
  Kuzuoka}{Vinson et~al\mbox{.}}{2018}]%
        {vinson2018human}
\bibfield{author}{\bibinfo{person}{David~W Vinson}, \bibinfo{person}{Leila
  Takayama}, \bibinfo{person}{Jodi Forlizzi}, \bibinfo{person}{Wendy Ju},
  \bibinfo{person}{Maya Cakmak}, {and} \bibinfo{person}{Hideaki Kuzuoka}.}
  \bibinfo{year}{2018}\natexlab{}.
\newblock \showarticletitle{Human-Robot Teaming}. In
  \bibinfo{booktitle}{\emph{Extended Abstracts of the 2018 CHI Conference on
  Human Factors in Computing Systems}}.
\newblock


\bibitem[\protect\citeauthoryear{Wallace and Boyd-Graber}{Wallace and
  Boyd-Graber}{2018}]%
        {wallace2018trick}
\bibfield{author}{\bibinfo{person}{Eric Wallace} {and} \bibinfo{person}{Jordan
  Boyd-Graber}.} \bibinfo{year}{2018}\natexlab{}.
\newblock \showarticletitle{Trick Me If You Can: Adversarial Writing of Trivia
  Challenge Questions}. In \bibinfo{booktitle}{\emph{Proceedings of ACL 2018
  Student Research Workshop}}.
\newblock


\end{thebibliography}

\end{document}